\documentclass{article}

\PassOptionsToPackage{numbers}{natbib}
\usepackage[main,final]{neurips_2026}

\usepackage[utf8]{inputenc} 
\usepackage[T1]{fontenc}    
\usepackage{hyperref}       
\usepackage{url}            
\usepackage{booktabs}       
\usepackage{amsfonts}       
\usepackage{nicefrac}       
\usepackage{microtype}      
\usepackage{xcolor}         

\usepackage{amsmath}
\usepackage{graphicx}
\usepackage{ulem}
\usepackage{multirow}
\usepackage{wrapfig}
\newtheorem{theorem}{Theorem}
\newtheorem{lemma}{Lemma}
\usepackage{algorithm}
\usepackage{algpseudocode}

\title{NDPP-Grasp: Non-Differentiable Physical Plausibility Constraint-Guided Task-Oriented Dexterous Grasp Generation}

\author{
  Qiuchi Xiang \\
  Lancaster University\\
  UK \\
  \texttt{q.xiang1@lancaster.ac.uk} \\
  \And
  Haoxuan Qu \\
  Lancaster University\\
  UK \\
  \texttt{h.qu5@lancaster.ac.uk} \\
  \AND
  Hossein Rahmani \\
  Lancaster University\\
  UK \\
  \texttt{h.rahmani@lancaster.ac.uk} \\
  \And
  Jun Liu\thanks{Corresponding Author.} \\
  Lancaster University\\
  UK \\
  \texttt{j.liu81@lancaster.ac.uk} \\
}

\begin{document}

\maketitle

\begin{abstract}
Task-oriented dexterous grasp generation aims to produce dexterous grasp poses that are both physically plausible and functionally suitable for specified manipulation tasks. Existing diffusion-based methods often address these two requirements in a decoupled manner: they first train a grasp diffusion model for task alignment and then rely on post-generation refinement to improve physical plausibility. However, this after-the-fact correction strategy applies physical plausibility guidance only once the grasp has already been generated, leaving the generation trajectory itself unguided by physical constraints and potentially leading to suboptimal grasps. To address this problem, we propose a novel framework that directly injects physical plausibility guidance into the denoising process of a task-aligned grasp diffusion model in a practical and effective manner, even when physical plausibility constraints are non-differentiable. This allows physical plausibility to shape grasp generation throughout denoising while preserving task alignment. Extensive experiments demonstrate the efficacy of our framework.

\end{abstract}

\section{Introduction}
\label{sec:intro}

\begin{wrapfigure}{r}{0.48\textwidth}
    \centering
    \vspace{-0.37cm}
    \includegraphics[width=0.47\textwidth]{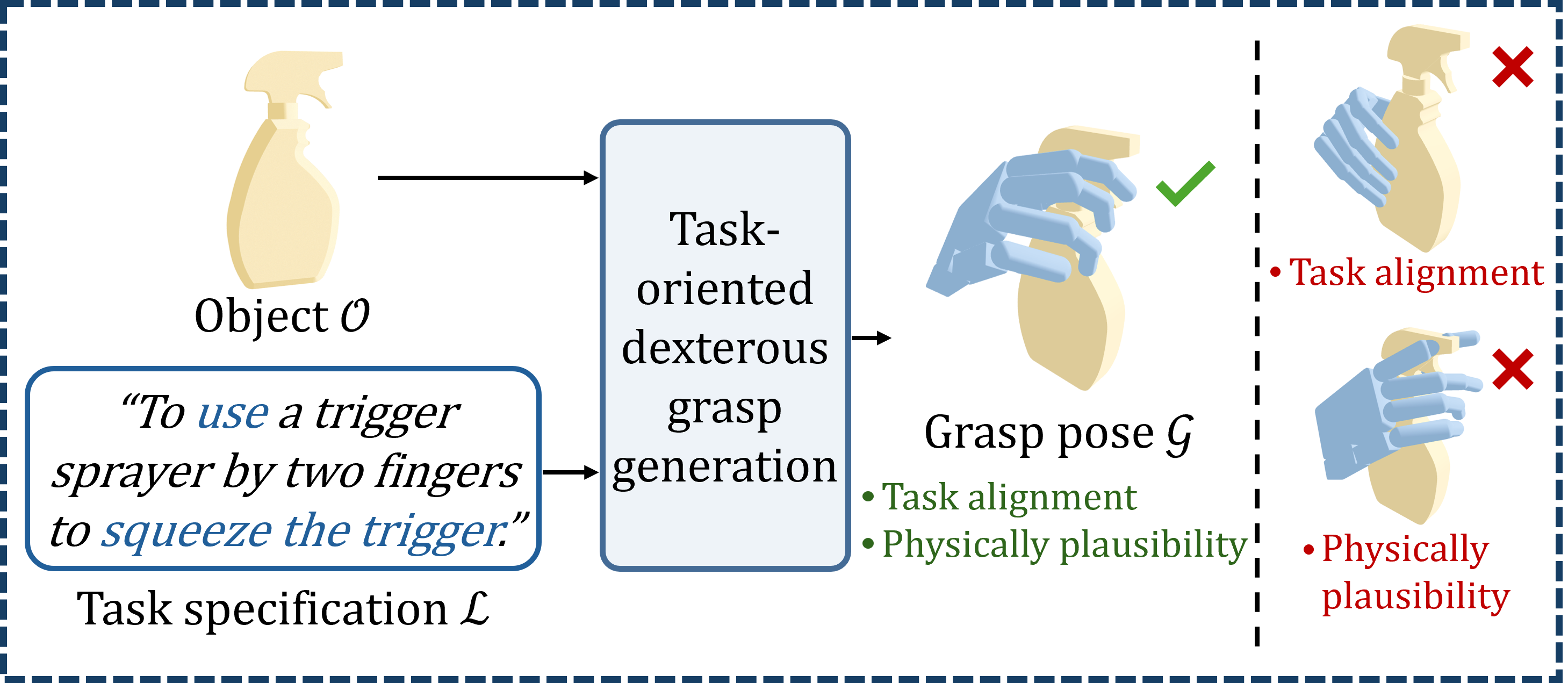}
    \vspace{-0.25cm}
    \caption{Illustration of task-oriented dexterous grasp generation.}
    \label{fig:infeasible_cases}
    \vspace{-0.4cm}
\end{wrapfigure}
Task-oriented dexterous grasp generation aims to synthesize dexterous hand poses for a given object and a specified manipulation task, such that the object can be grasped in a way both physically plausible and suitable for performing the task~\cite{jian2025gdexgrasp, wei2025afford, wei2024graspasyousay, zhang2024dextog}, as illustrated in Fig.~\ref{fig:infeasible_cases}. 
This problem is central to dexterous robotic manipulation because the grasp quality strongly affects whether the subsequent manipulation can be successfully executed \cite{wang2023dexgraspnet,fang2020learning}. 
Since dexterous robotic manipulation is useful in diverse practical scenarios, the significance of task-oriented dexterous grasp generation spans a wide range of applications~\cite{huang2025human, tsuji2025survey, xiangvision}, such as industrial manufacturing~\cite{zhang2025im_review}, household services~\cite{zhong2025anything}, and medical rehabilitation~\cite{huang2025human}. 
Owing to its significance, task-oriented dexterous grasp generation has recently received increasing attention \cite{zhang2024dextog, wei2024graspasyousay, chang2025text2grasp, wei2025afford}. In particular, motivated by strong generative capability of diffusion models, diffusion-based approaches have attracted growing interest as a promising direction for this problem~\cite{zhang2024dextog, wei2024graspasyousay, chang2025text2grasp}.

Despite recent progress, task-oriented dexterous grasp generation remains challenging because a practically useful grasp needs to satisfy two heterogeneous requirements simultaneously~\cite{jian2025gdexgrasp, jian2025zerodexgrasp}. Specifically, generated grasps need to align with manipulation intents in task specifications, while also being physically plausible in real-world interactions. These requirements are both essential. A grasp that satisfies the task specification may still be unusable if it is physically implausible, for example, if it penetrates the object.  Conversely, a physically plausible grasp may be functionally inappropriate for specified manipulation tasks. Therefore, effectively handling both requirements is central to generating practically useful grasps.

To handle this challenge, a seemingly straightforward strategy is to train a single generative model, such as a diffusion model, to satisfy both requirements jointly~\cite{zhu2023toward, chen2024task}. However, this strategy can be difficult to realize effectively in practice. On the one hand, incorporating the two heterogeneous requirements into a single training objective makes optimization challenging~\cite{wei2024graspasyousay, wei2025afford, jian2025gdexgrasp}. On the other hand, while alignment with the task specification can often be supervised in a differentiable manner, physical plausibility is more difficult to impose during training. Many physical plausibility criteria, such as hard hand-object non-penetration and self-collision checks, are inherently non-differentiable \cite{chen2025robust, paulusdifferentiable}. This makes them difficult to directly optimize and often requires differentiable surrogate losses instead~\cite{wei2024graspasyousay, wei2025afford, zhang2024dextog}. 
Although such surrogate objectives provide tractable training signals, the gap between the surrogate losses and the original physical plausibility constraints can limit the effectiveness of training~\cite{paulusdifferentiable, wei2024graspasyousay}. As a result, directly training a single model to satisfy both requirements often leads to suboptimal grasp generation performance~\cite{zhang2024dextog, wei2024graspasyousay, chang2025text2grasp}.

To avoid directly optimizing both requirements within a single generative model, existing methods~\cite{zhang2024dextog, wei2024graspasyousay, wei2025afford} often adopt a decoupled strategy. Instead of operating in a single process, they decompose task-oriented dexterous grasp generation into two stages. In the first stage, they focus on task alignment and train a generative model to produce a dexterous grasp that matches the task specification. In the second stage, they apply a refinement process to adjust the generated grasp and improve its physical plausibility~\cite{zhang2024dextog, wei2024graspasyousay}. 
However, although such decoupled two-stage methods can significantly improve task alignment by allowing the diffusion model to focus on matching the task specification, they still have clear limitations. First, since physical plausibility constraints are imposed only after generation, the generated grasp may have already drifted far into physically implausible regions during the denoising process, making it difficult for the refinement stage to recover a plausible grasp~\cite{jungjoint, liang2025simultaneous}. Second, in existing methods \cite{zhang2024dextog, wei2024graspasyousay}, the post-generation refinement stage still often relies on differentiable surrogate objectives for physical plausibility constraints~\cite{hasson2019learning, grady2021contactopt}, and thus remains limited by the mismatch with original non-differentiable physical plausibility criteria~\cite{paulusdifferentiable, wei2024graspasyousay}.

As a result, the above two strategies each lead to their own problems in task-oriented dexterous grasp generation. Hence, in this work, we aim to address these problems from a different perspective. Our key idea is to first train a grasp diffusion model for task alignment, as in existing decoupled two-stage methods~\cite{zhang2024dextog, chang2025text2grasp}. Then, instead of using a post-generation refinement stage, we inject guidance from physical plausibility constraints into the denoising process of the trained diffusion model. This design differs from both existing strategies. Unlike methods that impose task alignment and physical plausibility together during training~\cite{zhu2023toward, chen2024task}, our framework keeps models focused on task alignment and avoids mixing heterogeneous objectives in the diffusion training process, which can otherwise compromise training effectiveness~\cite{wei2024graspasyousay, wei2025afford}. Unlike post-generation refinement methods~\cite{zhang2024dextog, chang2025text2grasp}, physical plausibility is not applied only after a completely generated grasp. Instead, it progressively shapes the grasp throughout the denoising process. In this way, our framework can avoid mixing heterogeneous objectives while allowing physical plausibility to guide grasp generation before it is complete.

However, effectively realizing this idea remains challenging. Existing  techniques for guiding the grasp denoising process typically rely on differentiable objectives to construct guidance signals~\cite{zhong2025anything, 11127957}. Therefore, naively applying them to physical plausibility constraints would require replacing the original non-differentiable criteria with differentiable surrogate losses, leading to a mismatch of interest. To address this challenge, we propose \textbf{N}on-\textbf{D}ifferentiable \textbf{P}hysical \textbf{P}lausibility Constraint-Guided Task-Oriented Dexterous \textbf{Grasp} Generation (\textbf{NDPP-Grasp}), a novel diffusion-based framework for task-oriented dexterous grasp generation. To the best of our knowledge, NDPP-Grasp is the first framework that enables progressive denoising-time guidance from non-differentiable physical plausibility constraints for diffusion-based task-oriented dexterous grasp generation, without relying on differentiable surrogate losses. Below, we outline how NDPP-Grasp achieves this goal.

We first observe that progressively steering the denoising trajectory toward a final grasp that satisfies physical plausibility constraints can be viewed as a sequential decision-making problem under the denoising dynamics of grasp diffusion generation. At each denoising step, the guidance determines how to adjust the current grasp sample, so that the final denoised grasp can better satisfy the non-differentiable physical plausibility constraints. This viewpoint naturally connects our problem to stochastic optimal control~ \cite{yong1999stochastic, pandey2025variational,huang2024symbolic,hui2026training,zhang2022path}, which also studies how to make decisions over time to drive a stochastic dynamic system toward desired goals. Motivated by this connection, in our framework, we first explore whether stochastic control can help realize non-differentiable guidance during the denoising process. 
Through this exploration, we  observe that, by combining this viewpoint with the intrinsic structure of grasp diffusion generation, it can be theoretically shown that non-differentiable guidance during grasp denoising can indeed be achieved using only forward evaluations of physical plausibility constraints, without requiring these constraints to be differentiable.
This result provides a principled and effective way to directly apply non-differentiable physical plausibility constraints during grasp denoising, without relying on differentiable surrogate losses. However, while effective in principle, directly computing the derived formulation can be computationally intractable, thus impractical (see Sec.~\ref{sec:lookahead} for details).
To address this issue, we further exploit the structure of grasp diffusion generation and propose an amortized lookahead strategy in our framework. With this strategy, non-differentiable guidance during grasp denoising can be performed practically in real time. Together, these designs enable our framework to achieve practical and effective non-differentiable physical plausibility guidance during the denoising process of task-oriented dexterous grasp generation.

Our contributions are: (1) We propose NDPP-Grasp, a novel framework for task-oriented dexterous grasp generation. To the best of our knowledge, this is the first effort to enable non-differentiable physical plausibility guidance during the denoising process for task-oriented dexterous grasp generation, without relying on differentiable surrogate losses.
(2) To make such denoising-time guidance practical and effective, we introduce several theoretically grounded  design choices into our framework.
(3) NDPP-Grasp achieves superior performance on the evaluated benchmarks.

\section{Related Works}
\label{sec:related}

\noindent\textbf{Dexterous Grasp Generation.} Dexterous grasp generation is a core problem in dexterous robotic manipulation. Under this task, prior studies can be broadly categorized into two groups: task-agnostic~\cite{weng2024dexdiffuser,lu2024ugg,zhong2025anything,huang2023scenediffuser} and task-oriented~\cite{zhang2024dextog,wu2025partdextog,li2024semgrasp,wei2024graspasyousay}. The former aims to generate physically plausible grasp poses for a given object, while the latter aims to generate grasp poses that are both physically plausible and functionally suitable for the specified manipulation task. Since downstream manipulation often requires task-specific grasping behaviors, task-oriented dexterous grasp generation has recently attracted increasing attention \cite{wei2024graspasyousay,zhang2024dextog,chang2025text2grasp}. Specifically, with the emergence of diffusion models as powerful generative models~\cite{li2025longdiff,li2026diffgraph,xue2026hrdit}, diffusion-based methods have become a promising direction for this problem.  DexGYS~\cite{wei2024graspasyousay} cascades a diffusion model with transformer-based refinement model for task alignment and physical plausibility. 
DexTOG~\cite{zhang2024dextog} refines diffusion-generated grasps with a test-time collision handling module to reduce penetration. Text2Grasp~\cite{chang2025text2grasp} uses text-guided diffusion followed by hand-object contact optimization that improves physical plausibility.

Despite their different designs, these works share a common paradigm of improving physical plausibility through a post-generation refinement stage, where physical plausibility guidance is applied only after the grasp has already been generated. In contrast, we propose a novel framework that enables directly incorporating physical plausibility guidance into the denoising process of a learned diffusion-based dexterous grasp generation model in an effective and practical manner.

\noindent\textbf{Optimal Control.} Optimal control provides a classical perspective in control theory for studying how to choose actions over time to drive a dynamical system toward desired behaviors~\cite{naidu2018optimal, yong1999stochastic}. Recently, it has been studied in various areas~ \cite{domingoadjoint, rosolia2017autonomous, huang2024symbolic,hui2026training, bellicoso2018dynamic, liniger2015optimization, vanroye2023fatrop,zhang2022path}, such as robust locomotion~\cite{bellicoso2018dynamic}, autonomous racing~\cite{liniger2015optimization}, and trajectory optimization~\cite{vanroye2023fatrop}.  Inspired by these works, in this work, we propose a novel framework that leverages an optimal control perspective and further exploits the intrinsic structure of the grasp denoising process, collectively enabling physical plausibility guidance to be incorporated into the denoising process of a task-aligned grasp diffusion model in a practical and effective manner.

\section{Preliminaries}
\label{sec:preliminaries}

\textbf{Problem definition.} Let $\mathcal{O}$ denote an object representation, typically in point-cloud form, and let $\mathcal{L}$ denote a task specification, typically given as a natural-language sentence. Given $\mathcal{O}$ and $\mathcal{L}$, task-oriented dexterous grasp generation aims to synthesize a dexterous hand grasp pose $\mathcal{G} = (r, t, q)$, such that the object can be grasped by the hand in a way that is both physically plausible and suitable for performing the task. Here, $r \in SO(3)$ denotes the wrist rotation, $t \in \mathbb{R}^3$ denotes the wrist translation, and $q \in \mathcal{Q} \subseteq \mathbb{R}^J$ denotes the joint configuration of a $J$-DoF dexterous hand. The space $\mathcal{Q}$ denotes the valid joint configuration space defined by the hand kinematic limits.

\textbf{Diffusion-based task-oriented dexterous grasp generation.} We next describe the diffusion-based process for generating task-oriented dexterous grasps. Here, since the grasp pose $\mathcal{G}$ contains a wrist rotation $r$ in the non-Euclidean manifold $SO(3)$, directly applying standard Gaussian diffusion to generate $\mathcal{G}$ is not straightforward. Therefore, existing diffusion-based grasp generation methods~\cite{wei2024graspasyousay, zhang2024dextog} typically generate $\mathcal{G}$ through the following procedure. They first perform Gaussian diffusion in a Euclidean parameter space $\mathbb{R}^d$, progressively denoising Gaussian noise $\mathbf{y}_T \sim \mathcal{N}(0, I)$  into a target grasp representation $\mathbf{y}_0$ in this space, via a reverse-time stochastic differential equation (SDE). The generated representation $\mathbf{y}_0$ is then decoded into the grasp pose $\mathcal{G}$ through a mapping function $\Phi$, i.e., $\mathcal{G} = \Phi(\mathbf{y}_0)$. Specifically, the reverse-time SDE from $s=T$ to $s=0$ can be written as:
 \begin{equation}
\mathrm{d}\mathbf{y}_s = -\beta(s)\Big[\tfrac{1}{2}\mathbf{y}_s+\nabla_{\mathbf{y}_s}\log p_s(\mathbf{y}_s \mid \mathcal{O}, \mathcal{L}) \Big]\,\mathrm{d}s + \sqrt{\beta(s)}\,\mathrm{d}\bar{\mathbf{w}}_s,
\label{eq:reverse_sde}
\end{equation}
where $\beta(s)$ is the noise schedule, $\nabla_{\mathbf{y}_s}\log p_s(\mathbf{y}_s \mid \mathcal{O}, \mathcal{L})$ is conditional score function estimated by the diffusion model, $\mathrm{d}\bar{\mathbf{w}}_s$ is a Wiener process, and $\mathrm{d}s$ denotes a negative infinitesimal timestep. 

\section{Methodology}
\label{sec:methodology}

In task-oriented dexterous grasp generation, the goal is to generate grasps that are both task-aligned and physically plausible. Existing strategies address this challenge in different ways, but each has its own limitations. If physical plausibility constraints are added to the grasp diffusion training objective together with task alignment constraints, training can become complicated, and both task alignment and physical plausibility may be compromised~\cite{wei2024graspasyousay, wei2025afford}. If physical plausibility is instead improved by a separate post-generation refinement step, the guidance is applied only after grasp generation, which can limit its effectiveness~\cite{liang2025simultaneous, jungjoint}. In this work, we address this problem from a different perspective. We first train a grasp diffusion model to generate task-aligned grasps. Then, during inference, we inject physical plausibility guidance directly into the denoising process of the trained model. In this way, training can focus on task alignment, while physical plausibility is enforced during generation. 

However, making this perspective practically effective is non-trivial. Existing  techniques for guiding the grasp denoising process~\cite{zhong2025anything, 11127957} usually rely on differentiable objectives to provide guidance signals. Therefore, naively applying them to non-differentiable physical plausibility constraints would require replacing these constraints with differentiable surrogate losses, leading to a mismatch in optimization target. To address this challenge, we propose NDPP-Grasp, a novel and theoretically grounded framework for non-differentiable physical plausibility guidance in grasp diffusion.  It is built upon two key design choices. 
First,  motivated by previous approaches on optimal control and its connections to diffusion denoising process \cite{yong1999stochastic,huang2024symbolic,hui2026training,zhang2022path}, and exploiting the intrinsic structure of grasp diffusion denoising, we  adopt a similar approach to obtain a denoising-time guidance mechanism that only requires forward evaluations of physical plausibility constraints, without requiring the constraint evaluator to be differentiable (Sec.~\ref{sec:control}).  Second, although this guidance mechanism is effective in principle, directly computing it can be computationally intractable, as elaborated in Sec.~\ref{sec:lookahead}.
We therefore design an amortized lookahead strategy that makes the mechanism practically usable, enabling effective non-differentiable physical plausibility guidance while maintaining real-time grasp generation (Sec.~\ref{sec:lookahead}).

\subsection{Non-Differentiable Physical Plausibility Guidance for Grasp Denoising}
\label{sec:control}

In this section, we  introduce the non-differentiable denoising-time guidance mechanism  outlined above. Specifically, given a grasp diffusion model trained for task alignment, our goal is to guide its denoising process directly with non-differentiable physical plausibility constraints, without first converting them into differentiable surrogate losses. To derive such a guidance mechanism, below, we first specify what final grasps the guided denoising process should favor, which provides the target objective that the denoising-time guidance should optimize.

\noindent\textbf{Target objective formulation.}
A high-quality guided grasp should satisfy two requirements: it should remain task-aligned, and it should be physically plausible. For task alignment, since the diffusion model has been trained to generate task-aligned grasps conditioned on the object representation $\mathcal{O}$ and the task specification $\mathcal{L}$, preserving task alignment means keeping the final denoised grasp representation $\mathbf{y}_0$ close to the conditional distribution $p(\mathbf{y}_0 \mid \mathcal{O}, \mathcal{L})$ learned by the trained model. To encourage physical plausibility, let $c$ denote a physical plausibility constraint, and let $f_c$ denote its constraint evaluator, which measures the violation of $c$ for a grasp pose, with lower values indicating better constraint satisfaction. Following~\cite{chungdiffusion,song2023loss}, we model satisfaction of this constraint as an energy-based likelihood $p(c \mid \mathbf{y}_0) \propto \exp\big(-f_c(\mathbf{y}_0)\big)$. This likelihood assigns higher probability to grasps that better satisfy the physical plausibility constraint.

Combining the task-aligned generation distribution with this physical plausibility likelihood, the target objective can therefore be modeled as sampling $\mathbf{y}_0$ from the posterior
 \begin{equation}
p(\mathbf{y}_0 \mid \mathcal{O}, \mathcal{L}, c)
\;\propto\;
p(\mathbf{y}_0 \mid \mathcal{O}, \mathcal{L})\,p(c \mid \mathbf{y}_0)
\;=\;
p(\mathbf{y}_0 \mid \mathcal{O}, \mathcal{L})\,\exp\!\big(-f_c(\mathbf{y}_0)\big),
\label{eq:posterior}
\end{equation}
where $p(\mathbf{y}_0 \mid \mathcal{O}, \mathcal{L})$ preserves task alignment, while $\exp(-f_c(\mathbf{y}_0))$ enforces physical plausibility.

\noindent\textbf{Connection to stochastic optimal control.}
Having formulated the target posterior in Eq.~\ref{eq:posterior}, we now study how to guide the diffusion denoising process with non-differentiable physical plausibility constraints, so that the final samples follow this posterior.

To this end, we first draw inspiration from existing methods that guide denoising with differentiable objectives~\cite{zhong2025anything, 11127957}. These methods usually introduce an additional guidance term $\boldsymbol{\eta}_s(\mathbf{y}_s, s)$ into the reverse-time SDE in Eq.~\ref{eq:reverse_sde}, leading to the guided denoising dynamics:
 \begin{equation}
\mathrm{d}\mathbf{y}_s = -\beta(s)\Big[\tfrac{1}{2}\mathbf{y}_s + \nabla_{\mathbf{y}_s}\log p_s(\mathbf{y}_s \mid \mathcal{O}, \mathcal{L}) + \boldsymbol{\eta}_s(\mathbf{y}_s, s)\Big]\,\mathrm{d}s + \sqrt{\beta(s)}\,\mathrm{d}\bar{\mathbf{w}}_s,
\label{eq:guided_sde}
\end{equation}
In differentiable cases, $\boldsymbol{\eta}_s$ is typically constructed from the gradient of the constraint evaluator, e.g., $\boldsymbol{\eta}_s(\mathbf{y}_s, s)=-\nabla_{\mathbf{y}_s} f_c(\mathbf{y}_s)$, where we slightly abuse notation and let $f_c$ also denote the evaluator of a differentiable constraint. This observation provides a useful way to reformulate our problem. In Eq.~\ref{eq:guided_sde}, constraint guidance is encoded by the additional denoising drift term $\boldsymbol{\eta}_s$. In our non-differentiable setting, the key question, therefore, becomes how to construct a proper $\boldsymbol{\eta}_s$ without relying on such gradients. This reformulation gives us a concrete target: we need to find a guidance term that can play the role of constraint guidance in the denoising dynamics, even when the constraint evaluator itself is non-differentiable. However, the problem remains challenging: without analytic gradients, it is unclear how to design $\boldsymbol{\eta}_s$ so that the denoising process can be steered toward satisfying the non-differentiable physical plausibility constraints.

To seek a way to construct such a guidance term, we first examine Eq.~\ref{eq:guided_sde} more closely. We observe that $\boldsymbol{\eta}_s$ introduces an additional constraint-driven signal at each denoising step. This makes guided denoising resemble a sequential decision-making problem: at each step, one needs to decide how to guide the current sample so that the final grasp better satisfies the target constraints. This structure naturally connects to stochastic optimal control~ \cite{yong1999stochastic,huang2024symbolic,hui2026training,zhang2022path}, which similarly studies how to choose control signals over time to drive a stochastic dynamical system toward a desired objective.

Motivated by this connection, we  observe that an optimal control perspective can help construct $\boldsymbol{\eta}_s$ for non-differentiable physical plausibility constraints.  In particular, viewing $\boldsymbol{\eta}_s$ as a control signal allows  its construction to be formulated as a concrete optimization problem that aims to improve final constraint satisfaction under the denoising dynamics.  It can further be observed that, by leveraging the intrinsic structure of grasp diffusion denoising, this optimization problem admits an analytical solution. This solution yields a gradient-free guidance term that can be computed using only forward evaluations of the physical plausibility constraint evaluator, thereby enabling the desired non-differentiable denoising-time guidance mechanism. Below,  the formulation of this optimization problem and  its analytical solution are detailed.

\noindent\textbf{Optimization problem formulation.}
We now formulate the construction of the guidance term $\boldsymbol{\eta}_s$ as a concrete optimization problem from a stochastic optimal control perspective.

\uline{Step 1.}
We first rewrite Eq.~\ref{eq:guided_sde} into a standard control-system form. Since the reverse-time SDE is run from $s=T$ to $0$, while optimal control is usually in forward time, we introduce the forward-time denoising state
$\mathbf{g}_\tau = \mathbf{y}_{T-\tau}$,
where $\tau$ runs from $0$ to $T$. Under this change of variable, Eq.~\ref{eq:guided_sde} can be equivalently written as the following control-affine stochastic dynamical system (Appendix~\ref{sec:app_dynamics}):
 \begin{equation}
\mathrm{d}\mathbf{g}_\tau
=
\mathbf{F}(\mathbf{g}_\tau, \tau)\mathrm{d}\tau
+
\sigma(\tau)
\big(
\mathbf{u}_\tau(\mathbf{g}_\tau, \tau)\mathrm{d}\tau
+
\mathrm{d}\mathbf{w}_\tau
\big),
\label{eq:control_system}
\end{equation}
where
$\mathbf{u}_\tau(\mathbf{g}_\tau, \tau)
=
\sqrt{\beta(T{-}\tau)}
\boldsymbol{\eta}_{T-\tau}(\mathbf{g}_\tau, T{-}\tau)$
is the forward-time control signal induced by the denoising-time guidance term $\boldsymbol{\eta}_s$,
$\mathbf{F}(\mathbf{g}_\tau, \tau)
=
\frac{1}{2}\beta(T{-}\tau)\mathbf{g}_\tau
+
\beta(T{-}\tau)
\nabla_{\mathbf{g}_\tau}
\log p_{T-\tau}(\mathbf{g}_\tau \mid \mathcal{O}, \mathcal{L})$
is the uncontrolled drift,
$\sigma(\tau)=\sqrt{\beta(T{-}\tau)}$
is the diffusion coefficient, and $\mathrm{d}\mathbf{w}_\tau$ is a standard forward-time Wiener process. Here, $\mathbf{F}$ represents the original task-aligned denoising dynamics, while $\mathbf{u}_\tau$ represents the additional control signal for physical plausibility guidance.

Because $\mathbf{u}_\tau$ and $\boldsymbol{\eta}_s$ are related by a deterministic rescaling and time reversal, constructing $\boldsymbol{\eta}_s$ is equivalent to constructing $\mathbf{u}_\tau$. Therefore, in the remainder of this section, we work directly with $\mathbf{u}_\tau$.

\uline{Step 2.}
With this equivalence, the problem becomes finding an optimal control law $\mathbf{u}_\tau$ for the forward-time system in Eq.~\ref{eq:control_system}. Guided by the target posterior in Eq.~\ref{eq:posterior}, the control law should guide the denoising trajectory toward final grasps with better physical plausibility satisfaction, while keeping the controlled dynamics close to the original task-aligned denoising dynamics. Since $\mathbf{u}_\tau \equiv 0$ recovers the original denoising dynamics, this closeness can be encouraged by penalizing the magnitude of the control signal. This leads to the following stochastic optimal control objective:
 \begin{equation}
\mathbf{u}_{\tau:T}^* \;=\; \arg\min_{\mathbf{u}_{\tau:T}} \;
\mathbb{E}_{\mathcal{P}^{\mathbf{u}}}\!\left[\,\int_\tau^T \tfrac{1}{2}\|\mathbf{u}_\nu\|^2\,\mathrm{d}\nu \;+\; f_c(\mathbf{g}_T)\,\Big|\,\mathbf{g}_\tau\right],
\label{eq:cost}
\end{equation}
where $\mathbf{u}_{\tau:T}=\{\mathbf{u}_\nu\}_{\nu\in[\tau,T]}$ is the sequence of control signals applied from time $\tau$ to $T$, and $\mathcal{P}^{\mathbf{u}}$ is the controlled denoising dynamics induced by these control signals. 
The terminal cost $f_c(\mathbf{g}_T)$ penalizes physical plausibility violations of the final grasp, while the running cost $\frac{1}{2}\|\mathbf{u}_\nu\|^2$ penalizes the control magnitude to keep the guided dynamics close to the original task-aligned denoising dynamics.
This gives the concrete optimization problem for constructing the desired control signal.

\noindent\textbf{Derivation of the gradient-free optimal control law.}
Having formulated the optimization problem in Eq.~\ref{eq:cost},  its solution gives the optimal control law $\mathbf{u}_\tau^*$. Although this optimization problem appears difficult to solve in general,  it can be shown that, by leveraging the intrinsic structure of grasp diffusion denoising, including the Markov structure of denoising trajectories and the Gaussian noise structure,  the following analytical form of $\mathbf{u}_\tau^*$ can be obtained in our context:
 \begin{equation}
\mathbf{u}_\tau^*(\mathbf{g}_\tau, \tau)\,\mathrm{d}\tau
=
\frac{\mathbb{E}_{\mathcal{P}^0}\!\left[\exp\big(-f_c(\mathbf{g}_T)\big)\,\mathrm{d}\mathbf{w}_\tau \,\big|\,\mathbf{g}_\tau\right]}{\mathbb{E}_{\mathcal{P}^0}\!\left[\exp\big(-f_c(\mathbf{g}_T)\big) \,\big|\,\mathbf{g}_\tau\right]}.
\label{eq:forward_control}
\end{equation}
Here, $\mathcal{P}^0$ denotes the uncontrolled grasp denoising dynamics, i.e., the dynamics in Eq.~\ref{eq:control_system} with $\mathbf{u}_\tau \equiv 0$. The constraint evaluator $f_c$ measures the physical plausibility violation of final denoised grasp $\mathbf{g}_T$.

Eq.~\ref{eq:forward_control} is the key result of this section. It gives the optimal forward-time control signal for constraint-guided grasp denoising. Since $\mathbf{u}_\tau$ is deterministically related to the original guidance term $\boldsymbol{\eta}_s$ through the change of variables in Eq.~\ref{eq:control_system}, solving for $\mathbf{u}_\tau^*$ equivalently provides the desired denoising guidance. Importantly, Eq.~\ref{eq:forward_control} only requires forward evaluations of $f_c$ along uncontrolled denoising trajectories without differentiating $f_c$. Thus,  this control law enables non-differentiable physical plausibility guidance for grasp denoising without requiring differentiability or gradient computation.

\subsection{Practical Instantiation via Amortized Lookahead}
\label{sec:lookahead}

Above, we  obtain a principled form of the control law $\mathbf{u}_\tau^*$ that enables non-differentiable physical plausibility guidance for grasp denoising, without requiring differentiability or gradient computation.

\noindent\textbf{Remaining practical challenge.}
Yet, this principled control law cannot be directly computed in practice. As shown in Eq.~\ref{eq:forward_control}, $\mathbf{u}_\tau^*$ is expressed as an expectation under the uncontrolled grasp denoising dynamics $\mathcal{P}^0$, which is intractable to compute in closed form. Therefore, to use this control law as a practical guidance algorithm, we need an effective way to instantiate this expectation.

\noindent\textbf{A natural finite-lookahead instantiation.}
Since Eq.~\ref{eq:forward_control} involves an expectation under the uncontrolled denoising dynamics, a direct way to instantiate it is Monte Carlo sampling~\cite{robert2004monte}. Specifically, at each denoising step $\tau$, we sample $M$ uncontrolled future denoising trajectories. Then, for each sampled trajectory, we evaluate the physical plausibility of its final grasp. Thus, the key quantity we need from the $m$-th trajectory is its terminal grasp $\mathbf{g}_{T}^{(m)}$, which allows us to compute the terminal physical plausibility cost $f_c(\mathbf{g}_{T}^{(m)})$. With these sampled trajectories and their terminal costs, the expectation in Eq.~\ref{eq:forward_control} can be instantiated by Monte Carlo sampling. See Appendix~\ref{sec:app_mc_natural} for more details.

This reduces the practical problem of instantiating $\mathbf{u}_\tau^*$ to the problem of obtaining terminal grasps $\mathbf{g}_{T}^{(m)}$ for the sampled future trajectories. In principle, this can be achieved by simulating the uncontrolled denoising dynamics step by step, i.e.,
$\mathbf{g}_{\tau} \rightarrow \mathbf{g}_{\tau+1}^{(m)} \rightarrow \cdots \rightarrow \mathbf{g}_{T}^{(m)}$.
However, this requires simulating the entire remaining denoising process for every sampled trajectory. 
Since the guidance term needs to be instantiated at every denoising step, and each instantiation requires $M$ trajectories, this direct simulation introduces a substantial computational burden. This is particularly undesirable for dexterous grasp generation, which is often used in robotic manipulation pipelines where real-time inference is important~\cite{weimathcal}.

To tackle this problem, following the common use of terminal-data prediction in diffusion guidance~\cite{song2023loss, shen2024understanding}, a more practical alternative is to use Tweedie-based finite lookahead~\cite{efron2011tweedie}. Specifically, instead of simulating each trajectory all the way to the terminal step, we only simulate $H$ future denoising steps, where $H$ is a hyperparameter, i.e.,
$\mathbf{g}_{\tau} \rightarrow \mathbf{g}_{\tau+1}^{(m)} \rightarrow \cdots \rightarrow \mathbf{g}_{\tau+H}^{(m)}$.
Then, we use Tweedie's formula~\cite{efron2011tweedie} to predict the terminal grasp from the lookahead endpoint:
$\hat{\mathbf{g}}_{T}^{(m)}=\mathcal{T}_{\tau+H}(\mathbf{g}_{\tau+H}^{(m)})$,
where $\mathcal{T}_{\tau+H}(\cdot)$ denotes the Tweedie prediction operator at time $\tau+H$, and $\hat{\mathbf{g}}_{T}^{(m)}$ denotes the predicted terminal grasp. We then evaluate $f_c(\hat{\mathbf{g}}_{T}^{(m)})$ and use it in the Monte Carlo instantiation of Eq.~\ref{eq:forward_control}. In this way, we obtain a natural finite-lookahead instantiation of Eq.~\ref{eq:forward_control}: it keeps the Monte Carlo structure of the control-law instantiation, but replaces expensive full future trajectory simulation with Tweedie-based terminal-grasp prediction from $H$-step lookahead trajectories.

\noindent\textbf{Limitation of this natural solution.}
Although the natural finite-lookahead instantiation is more practical than simulating the full future trajectory, it still suffers from a key tension between effectiveness and practicality. Specifically, to obtain a reliable terminal grasp prediction, the lookahead horizon $H$ is usually expected to be sufficiently large, so that the terminal grasp prediction can rely on more future denoising context when applying Tweedie's formula. However, a larger $H$ also means that, at every denoising step, each sampled trajectory still needs to be simulated from $\mathbf{g}_{\tau}$ to $\mathbf{g}_{\tau+H}$. Since this process is repeated for $M$ trajectories at every denoising step, the instantiation can still be costly during sampling. In contrast, choosing a smaller $H$ reduces computation, but makes the terminal grasp prediction rely more heavily on noisier intermediate states and can weaken guidance effectiveness. Therefore, the above natural finite-lookahead instantiation still does not make Eq.~\ref{eq:forward_control} both practical and effective in a satisfactory way.

\noindent\textbf{Our strategy.}
To address this issue, we propose an amortized lookahead strategy. The key idea behind our strategy is that consecutive $H$-step lookahead instantiations have highly overlapping future time ranges. Specifically, at denoising step $\tau$, the natural finite-lookahead instantiation samples future states
$(\mathbf{g}_{\tau+1}^{(m)},\ldots,\mathbf{g}_{\tau+H}^{(m)})$
for each trajectory. At the next denoising step $\tau+1$, if we performed a fresh instantiation from scratch, we would need to sample new future states from the actual current state $\mathbf{g}_{\tau+1}$, namely
$(\tilde{\mathbf{g}}_{\tau+2}^{(m)},\ldots,\tilde{\mathbf{g}}_{\tau+H+1}^{(m)})$.
These two lookahead instantiations therefore cover same future time indices from $\tau+2$ to $\tau+H$. This motivates us to ask whether, at denoising step $\tau+1$, we can reuse the already generated states
$(\mathbf{g}_{\tau+2}^{(m)},\ldots,\mathbf{g}_{\tau+H}^{(m)})$
and only add one new transition to obtain $\mathbf{g}_{\tau+H+1}^{(m)}$. If this reuse is valid, then even with a large lookahead horizon $H$, each subsequent denoising step would only need to add one new transition per trajectory, instead of generating a complete $H$-step trajectory from scratch. This would directly address the tension above: we could preserve the effectiveness of a larger lookahead horizon while keeping per-step computation practical.

However, such reuse is not directly valid. The stored future states from step $\tau$ were generated after passing through the sampled candidate state $\mathbf{g}_{\tau+1}^{(m)}$, whereas the fresh instantiation at step $\tau+1$ should start from the actual current state $\mathbf{g}_{\tau+1}$. Since these two starting states are different, directly treating the stored future states as freshly sampled ones would correspond to using an incorrect future trajectory distribution. Fortunately, this mismatch can be corrected in our context. Specifically, by using the Markov property of the uncontrolled grasp denoising dynamics and the Gaussian form of its one-step transition, we show that the mismatch between the desired fresh future segment
$(\tilde{\mathbf{g}}_{\tau+2}^{(m)},\ldots,\tilde{\mathbf{g}}_{\tau+H}^{(m)})$
and the reused stored future segment
$(\mathbf{g}_{\tau+2}^{(m)},\ldots,\mathbf{g}_{\tau+H}^{(m)})$
can be captured by a closed-form correction weight. Formally, we show that (derivation provided in Appendix~\ref{sec:app_amortized}):
\begin{equation}
\begin{aligned}
\mathbb{E}_{\mathcal{P}^0_{\tau+2:\tau+H}(\cdot \mid \mathbf{g}_{\tau+1})}
\left[
\Psi(\tilde{\mathbf{g}}_{\tau+2}^{(m)},\ldots,\tilde{\mathbf{g}}_{\tau+H}^{(m)})
\right] =
\mathbb{E}_{\mathcal{P}^0_{\tau+2:\tau+H}(\cdot \mid \mathbf{g}_{\tau+1}^{(m)})}
\left[
\rho_{\tau+1}^{(m)}
\,
\Psi(\mathbf{g}_{\tau+2}^{(m)},\ldots,\mathbf{g}_{\tau+H}^{(m)})
\right].
\end{aligned}
\label{eq:reuse_correction}
\end{equation}
Here, $\Psi$ denotes a trajectory functional, i.e., a function applied to a lookahead trajectory to compute the quantities needed in the finite-lookahead Monte Carlo instantiation, see Appendix~\ref{sec:app_lookahead} for details. The distribution $\mathcal{P}^0_{\tau+2:\tau+H}(\cdot \mid \cdot)$ denotes the conditional distribution of the uncontrolled future states from time $\tau+2$ to $\tau+H$, given the state at time $\tau+1$. The scalar $\rho_{\tau+1}^{(m)}$ is the correction weight for reusing the stored future states, whose closed-form expression is derived in Appendix~\ref{sec:app_amortized}.

Based on this result, we instantiate Eq.~\ref{eq:forward_control} through an amortized lookahead procedure. The key idea is to reuse stored future states across consecutive denoising steps with correction weights, and only append one new transition per step to maintain the lookahead horizon. The procedure consists of five operations: the first is performed at the initial denoising step, while the remaining four are repeated at each subsequent denoising step. Below, we detail these five operations.
\uline{(1) Initialization.} At the initial denoising step, since no previous lookahead trajectories are available for reuse, we perform the natural finite-lookahead instantiation once and store the resulting $M$ lookahead trajectories of length $H$ for later reuse. 
\uline{(2) Reuse with correction.} At a subsequent denoising step $\tau+1$, for each stored trajectory, we reuse the stored future states
$(\mathbf{g}_{\tau+2}^{(m)},\ldots,\mathbf{g}_{\tau+H}^{(m)})$
instead of simulating them again from scratch. Since these states were generated from the stored candidate state $\mathbf{g}_{\tau+1}^{(m)}$ rather than the actual current state $\mathbf{g}_{\tau+1}$, we compute the correction weight $\rho_{\tau+1}^{(m)}$ and use it to correct the corresponding Monte Carlo contribution. 
\uline{(3) Extension.} To maintain an $H$-step lookahead horizon at denoising step $\tau+1$, we append one additional uncontrolled transition from $\mathbf{g}_{\tau+H}^{(m)}$ to $\mathbf{g}_{\tau+H+1}^{(m)}$ using the original uncontrolled denoising dynamics.
\uline{(4) Monte Carlo computation.} After reuse and extension, we use the resulting future segment
$(\mathbf{g}_{\tau+2}^{(m)},\ldots,\mathbf{g}_{\tau+H}^{(m)},\mathbf{g}_{\tau+H+1}^{(m)})$
to perform the finite-lookahead Monte Carlo computation at the current state $\mathbf{g}_{\tau+1}$, with its contribution multiplied by $\rho_{\tau+1}^{(m)}$. This gives the current-step Monte Carlo instantiation of the expectation in Eq.~\ref{eq:forward_control}, see Appendix~\ref{sec:app_mc_amortized} for more details. 
\uline{(5) Storage for the next denoising step.} After the current guidance term is computed, we store the extended future trajectory
$(\mathbf{g}_{\tau+2}^{(m)},\ldots,\mathbf{g}_{\tau+H+1}^{(m)})$
for reuse in the next denoising step.

Through these five operations, the amortized procedure avoids regenerating a complete $H$-step lookahead trajectory from scratch at every denoising step. Thus, we can use a sufficiently large lookahead horizon $H$ to improve the reliability of terminal grasp prediction, while after initialization, each denoising step only needs to simulate one additional transition per trajectory. This turns Eq.~\ref{eq:forward_control} into a practical and effective algorithm for computing non-differentiable physical plausibility guidance during grasp denoising. 

\section{Experiments}
\label{sec:experiments}

\noindent\textbf{Datasets \& Evaluation Metrics.} We evaluate NDPP-Grasp on two widely used datasets: DexGYSNet~\cite{wei2024graspasyousay} and DexTOG-80K~\cite{zhang2024dextog}. On DexGYSNet, we use the standard train-test split~\cite{wei2024graspasyousay} and additionally evaluate on an open-set split. Following~\cite{wei2024graspasyousay}, we evaluate physical quality using success rate (Succ.), $Q_1$, and maximal penetration depth (Pen.), evaluate task alignment using Fréchet Inception Distance (FID), Chamfer Distance (CD), and Contact Distance (Con.), and evaluate diversity using the standard deviations of translation $\delta_t$, rotation $\delta_r$, and joint angles $\delta_q$. On DexTOG-80K, we use its standard split. Following~\cite{zhang2024dextog}, we report the overall success rate (Overall-Succ.), which captures both physical quality and task alignment. We also follow \cite{zhang2024dextog} to report three metrics focused on physical plausibility: $Q_1$, Pen., and the collision-free rate $\eta_f$. On both datasets, we also report the average per-grasp inference time, denoted as ``Time'' in the tables. See Appendix~\ref{sec:app_experiments} for more details.

\noindent\textbf{Experimental Setups \& Implementation Details.} Across the two datasets, we apply our method to grasp diffusion models trained following the respective training procedures of three recent diffusion-based task-oriented dexterous grasp generation works: Text2Grasp~\cite{chang2025text2grasp}, DexTOG~\cite{zhang2024dextog}, and DexGYS~\cite{wei2024graspasyousay}. Following~\cite{wei2024graspasyousay, wei2025afford}, we consider three non-differentiable physical plausibility constraints: the hand-object non-penetration constraint, self-collision avoidance, and the joint-limit constraint. We set hyperparameters $M = 8$ and $H = 20$. All experiments are conducted on  NVIDIA A6000 GPU. See Appendix~\ref{sec:app_experiments} for more experimental setups and implementation details.

\subsection{Main Results}
\label{sec:main_results}

In Tab.~\ref{tab:dexgysnet} and Tab.~\ref{tab:dextog}, we report results on DexGYSNet and DexTOG-80K, respectively. As shown, when using the same grasp diffusion model, our method consistently achieves substantial improvements in physical plausibility compared with existing methods that enhance physical plausibility through various post-generation refinement strategies, demonstrating the efficacy of our approach. When compared with other recent task-oriented dexterous grasp generation methods, our method also achieves superior performance, further demonstrating its effectiveness.

\begin{table}[h]
\centering
\caption{Performance comparison on DexGYSNet. ``Ours (training following X)'' denotes our method applied to the grasp diffusion model trained following the training procedure of X (see Appendix~\ref{sec:app_algorithm}).}
\resizebox{1\textwidth}{!}{
\begin{tabular}{l|ccc|ccc|ccc|ccc|ccc|ccc|c}
\toprule
\multirow{3}{*}{Method} & \multicolumn{9}{c|}{Standard train-test split on the DexGYSNet dataset} & \multicolumn{9}{c|}{Open-set train-test split on the DexGYSNet dataset} & \multirow{3}{*}{\shortstack{Time \\(ms) $\downarrow$}} \\
\cmidrule(lr){2-10} \cmidrule(lr){11-19}
 & \multicolumn{3}{c|}{\textbf{Physical Qual.}} & \multicolumn{3}{c|}{Task Align.} & \multicolumn{3}{c|}{Diversity} & \multicolumn{3}{c|}{\textbf{Physical Qual.}} & \multicolumn{3}{c|}{Task Align.} & \multicolumn{3}{c|}{Diversity} & \\
 & Succ. $\uparrow$ & $Q_1$ $\uparrow$ & Pen. $\downarrow$ & FID $\downarrow$ & CD $\downarrow$ & Con. $\downarrow$ & $\delta_t$ $\uparrow$ & $\delta_r$ $\uparrow$ & $\delta_q$ $\uparrow$ & Succ. $\uparrow$ & $Q_1$ $\uparrow$ & Pen. $\downarrow$ & FID $\downarrow$ & CD $\downarrow$ & Con. $\downarrow$ & $\delta_t$ $\uparrow$ & $\delta_r$ $\uparrow$ & $\delta_q$ $\uparrow$ & \\
\midrule
\midrule
GraspCVAE~\cite{sohn2015learning} & 29.12\% & 0.054 & 0.551 & 31.26 & 3.138 & 0.096 & 0.179 & 1.762 & 0.179 & 24.36\% & 0.048 & 0.716 & 41.57 & 3.985 & 0.128 & 0.158 & 1.583 & 0.157 & 48.3 \\
GraspTTA~\cite{jiang2021hand} & 43.46\% & 0.071 & 0.188 & 35.41 & 12.19 & 0.111 & 2.111 & 6.150 & 3.869 & 38.71\% & 0.064 & 0.246 & 47.10 & 15.48 & 0.148 & 1.876 & 5.527 & 3.438 & 1873.4 \\
DGTR~\cite{xu2024dexgrasptrans} & 51.91\% & 0.078 & 0.163 & 23.31 & 2.895 & 0.078 & 2.037 & 14.01 & 4.299 & 46.23\% & 0.071 & 0.215 & 31.12 & 3.684 & 0.104 & 1.812 & 12.47 & 3.821 & 41.7 \\
SceneDiffuser~\cite{huang2023scenediffuser} & 62.24\% & 0.083 & 0.253 & 20.44 & 1.679 & 0.045 & 0.346 & 3.455 & 0.387 & 57.48\% & 0.076 & 0.330 & 27.26 & 2.139 & 0.060 & 0.306 & 3.117 & 0.343 & 44.1 \\
AffordDexGrasp~\cite{wei2025afford} & 62.81\% & 0.082 & 0.228 & 6.538 & 1.198 & 0.039 & 5.924 & 53.86 & 5.918 & 59.17\% & 0.077 & 0.279 & 8.712 & 1.523 & 0.047 & 5.307 & 48.58 & 5.368 & 493.1 \\
\midrule
Text2Grasp~\cite{chang2025text2grasp} & 60.43\% & 0.080 & 0.197 & 21.83 & 1.812 & 0.051 & 0.398 & 4.123 & 0.452 & 55.36\% & 0.073 & 0.272 & 28.43 & 2.301 & 0.065 & 0.357 & 3.642 & 0.405 & 287.4 \\
\textbf{Ours} (training following~\cite{chang2025text2grasp}) & \textbf{63.27\%} & \textbf{0.084} & \textbf{0.163} & \textbf{21.78} & \textbf{1.798} & \textbf{0.049} & \textbf{0.402} & \textbf{4.131} & \textbf{0.461} & \textbf{58.18\%} & \textbf{0.077} & \textbf{0.224} & \textbf{28.37} & \textbf{2.287} & \textbf{0.063} & \textbf{0.361} & \textbf{3.651} & \textbf{0.413} & \textbf{20.5} \\
\midrule
DexTOG~\cite{zhang2024dextog} & 62.67\% & 0.082 & 0.234 & 13.82 & 1.452 & 0.041 & 3.827 & 32.15 & 3.912 & 57.82\% & 0.075 & 0.253 & 18.43 & 1.849 & 0.054 & 3.394 & 28.97 & 3.472 & 389.6 \\
\textbf{Ours} (training following~\cite{zhang2024dextog}) & \textbf{65.94\%} & \textbf{0.086} & \textbf{0.192} & \textbf{13.79} & \textbf{1.438} & \textbf{0.040} & \textbf{3.842} & \textbf{32.23} & \textbf{3.931} & \textbf{61.08\%} & \textbf{0.079} & \textbf{0.213} & \textbf{18.27} & \textbf{1.836} & \textbf{0.052} & \textbf{3.412} & \textbf{29.04} & \textbf{3.489} & \textbf{17.5} \\
\midrule
DexGYS~\cite{wei2024graspasyousay} & 63.31\% & 0.083 & 0.223 & 6.538 & 1.198 & 0.036 & 6.118 & 55.68 & 6.118 & 58.42\% & 0.076 & 0.291 & 8.724 & 1.526 & 0.048 & 5.383 & 48.95 & 5.392 & 23.2 \\
\textbf{Ours} (training following~\cite{wei2024graspasyousay}) & \textbf{67.05\%} & \textbf{0.088} & \textbf{0.182} & \textbf{6.487} & \textbf{1.184} & \textbf{0.035} & \textbf{6.132} & \textbf{55.74} & \textbf{6.127} & \textbf{61.89\%} & \textbf{0.081} & \textbf{0.239} & \textbf{8.683} & \textbf{1.519} & \textbf{0.047} & \textbf{5.391} & \textbf{49.12} & \textbf{5.401} & \textbf{11.6} \\
\bottomrule
\end{tabular}
}
\label{tab:dexgysnet}
\end{table}

\begin{table}[htbp]
\centering
\caption{Performance comparison on DexTOG-80K.}

\resizebox{0.6\textwidth}{!}{
\begin{tabular}{l|c|c|c|c|c}
\toprule
Method & Overall-Succ. $\uparrow$ & $Q_1$ $\uparrow$ & Pen. $\downarrow$ & $\eta_f$ $\uparrow$ & Time (ms) $\downarrow$ \\
\midrule
\midrule
GraspTTA~\cite{jiang2021hand} & 3.20\% & 0.031 & 0.524 & 18.7\% & 1901.7 \\
AffordDexGrasp~\cite{wei2025afford} & 22.41\% & 0.076 & 0.462 & 35.2\% & 501.3 \\
\midrule
Text2Grasp~\cite{chang2025text2grasp} & 17.85\% & 0.062 & 0.391 & 28.7\% & 308.4 \\
\textbf{Ours} (training following~\cite{chang2025text2grasp}) & \textbf{20.41\%} & \textbf{0.067} & \textbf{0.323} & \textbf{34.2\%} & \textbf{20.7} \\
\midrule
DexGYS~\cite{wei2024graspasyousay} & 23.87\% & 0.078 & 0.448 & 36.8\% & 23.6 \\
\textbf{Ours} (training following~\cite{wei2024graspasyousay}) & \textbf{26.53\%} & \textbf{0.083} & \textbf{0.374} & \textbf{42.4\%} & \textbf{11.8} \\
\midrule
DexTOG~\cite{zhang2024dextog} & 28.60\% & 0.093 & 0.385 & 50.5\% & 395.8 \\
\textbf{Ours} (training following~\cite{zhang2024dextog}) & \textbf{31.92\%} & \textbf{0.098} & \textbf{0.318} & \textbf{56.8\%} & \textbf{17.7} \\
\bottomrule
\end{tabular}
}
\label{tab:dextog}
\end{table}

\subsection{Further Analysis and Ablation Studies}
\label{sec:ablation}

We conduct extensive analyses and ablation studies on the standard train-test split of DexGYSNet, using the grasp diffusion model trained following DexGYS~\cite{wei2024graspasyousay}. \textbf{More analyses and ablation studies are provided in Appendix~\ref{sec:app_ablation}.}

\noindent\textbf{Impact of gradient-free optimal control law.}
To evaluate the effectiveness of  the optimal control law, we compare against the following four variants (details in Appendix~\ref{sec:app_ablation}).
 \textbf{Variant A} uses surrogate losses of physical plausibility constraints during training.
\textbf{Variant B} uses the gradients of surrogates to guide the denoising process at inference time.
\textbf{Variant C} estimates gradients of the non-differentiable constraints via zeroth-order finite-difference approximation~\cite{spall2002multivariate}.
\textbf{Variant D} applies reinforcement learning~\cite{williams1992simple} to approximate gradient guidance for non-differentiable constraints.
As shown in Tab.~\ref{tab:ablation_control}, NDPP-Grasp outperforms all variants, demonstrating the effectiveness of  the gradient-free optimal control law.

\begin{table}[h]
\centering
\caption{Evaluation of  the optimal control law.}
\resizebox{0.60\textwidth}{!}{
\begin{tabular}{l|ccc|ccc|ccc}
\toprule
\multirow{2}{*}{Method} & \multicolumn{3}{c|}{\textbf{Physical Qual.}} & \multicolumn{3}{c|}{Task Align.} & \multicolumn{3}{c}{Diversity} \\
 & Succ. $\uparrow$ & $Q_1$ $\uparrow$ & Pen. $\downarrow$ & FID $\downarrow$ & CD $\downarrow$ & Con. $\downarrow$ & $\delta_t$ $\uparrow$ & $\delta_r$ $\uparrow$ & $\delta_q$ $\uparrow$ \\
\midrule
Variant A & 54.27\% & 0.071 & 0.378 & 12.64 & 1.987 & 0.061 & 4.213 & 41.28 & 4.317 \\
Variant B & 62.91\% & 0.084 & 0.211 & 6.583 & 1.192 & 0.036 & 6.094 & 55.21 & 6.092 \\
Variant C & 60.74\% & 0.079 & 0.238 & 6.728 & 1.203 & 0.038 & 6.024 & 54.41 & 6.038 \\
Variant D & 59.86\% & 0.077 & 0.247 & 6.793 & 1.214 & 0.039 & 5.978 & 53.97 & 5.991 \\
\midrule
Ours & \textbf{67.05\%} & \textbf{0.088} & \textbf{0.182} & \textbf{6.487} & \textbf{1.184} & \textbf{0.035} & \textbf{6.132} & \textbf{55.74} & \textbf{6.127} \\
\bottomrule
\end{tabular}
}
\label{tab:ablation_control}
\end{table}

\noindent\textbf{Impact of the amortized lookahead strategy.}
To evaluate the effectiveness of our amortized lookahead strategy, we compare two variants. In the (\textbf{Complete-trajectory sampling}), at each denoising step, we sample complete trajectories that denoise until the terminal step, to estimate the optimal control signal. In the (\textbf{$H$-step trajectory resampling}), we only sample $H$-step trajectories and apply Tweedie's formula to estimate. As shown in Tab.~\ref{tab:ablation_amortized}, our method achieves comparable performance with substantially lower computational cost, showing the practical effectiveness of our amortized lookahead strategy.

\begin{table}[h]
\centering
\caption{Evaluation of the amortized lookahead.}
\label{tab:ablation_amortized}
\resizebox{0.60\textwidth}{!}{
\begin{tabular}{l|ccc|ccc|ccc|c}
\toprule
\multirow{2}{*}{Method} & \multicolumn{3}{c|}{Physical Qual.} & \multicolumn{3}{c|}{Task Align.} & \multicolumn{3}{c|}{Diversity} & \multirow{2}{*}{\shortstack{Time \\(ms) $\downarrow$}} \\
 & Succ. $\uparrow$ & $Q_1$ $\uparrow$ & Pen. $\downarrow$ & FID $\downarrow$ & CD $\downarrow$ & Con. $\downarrow$ & $\delta_t$ $\uparrow$ & $\delta_r$ $\uparrow$ & $\delta_q$ $\uparrow$ & \\
\midrule
Complete-trajectory sampling & 67.06\% & 0.089 & 0.183 & 6.484 & 1.183 & 0.034 & 6.135 & 55.78 & 6.130 & 261.4 \\
$H$-step trajectory resampling & 67.05\% & 0.088 & 0.182 & 6.488 & 1.184 & 0.035 & 6.131 & 55.73 & 6.126 & 109.8 \\
\midrule
Ours & \textbf{67.05\%} & \textbf{0.088} & \textbf{0.182} & \textbf{6.487} & \textbf{1.184} & \textbf{0.035} & \textbf{6.132} & \textbf{55.74} & \textbf{6.127} & \textbf{11.6} \\
\bottomrule
\end{tabular}
}
\end{table}

\begin{figure}[h]
\centering
    \includegraphics[width=0.7\textwidth]{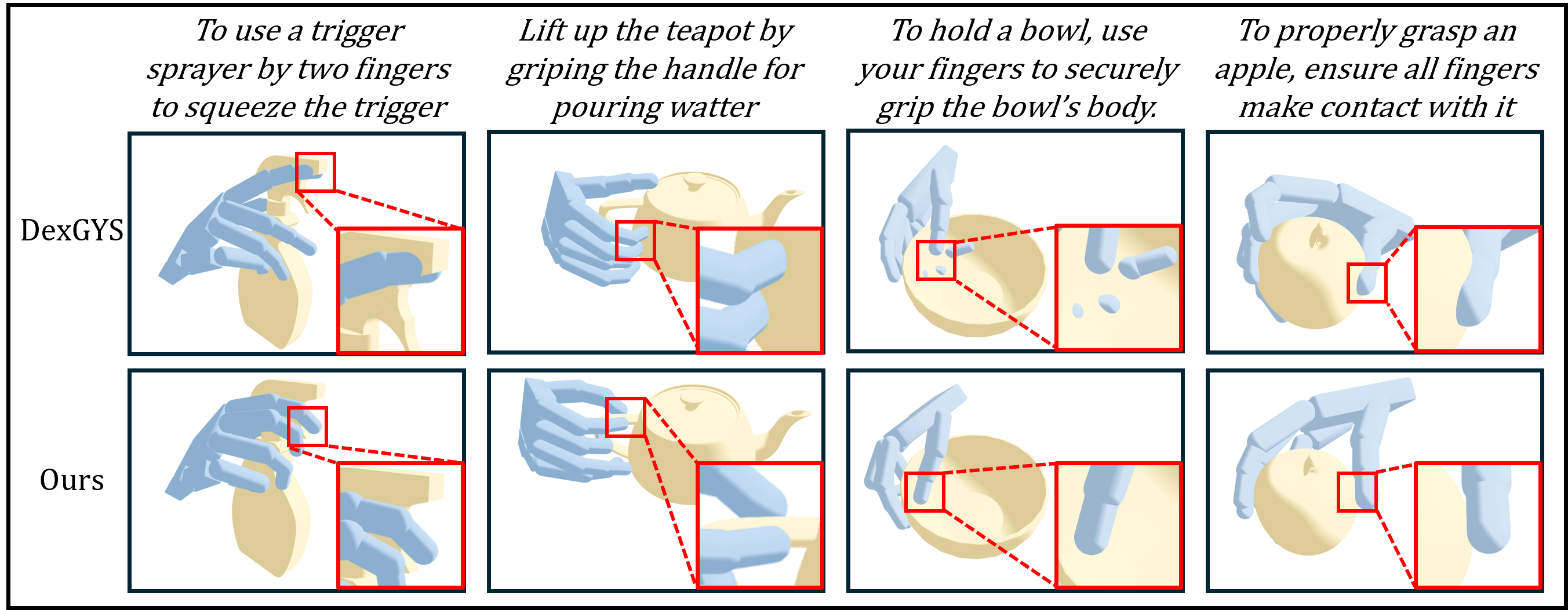}
\caption{Qualitative comparisons on DexGYSNet, with the key contact regions zoomed in \textcolor{red}{red boxes}.}
\label{fig:qualitative}
\end{figure}

\noindent\textbf{Qualitative Results.}
In addition to quantitative comparisons, we also provide qualitative comparisons in Fig.~\ref{fig:qualitative}. As shown, our framework produces grasps with strong physical plausibility while preserving alignment with the task specifications, whereas the compared method DexGYS~\cite{wei2024graspasyousay} produces physically implausible grasps that fail to satisfy the physical constraints, further showing the efficacy of our method. \textbf{More qualitative results and comparisons with more methods are in Appendix~\ref{sec:app_qualitative}.}

\section{Conclusion}
\label{sec:conclusions}
In this paper, we have proposed NDPP-Grasp, a novel and theoretically grounded framework for task-oriented dexterous grasp generation that enables practical and effective non-differentiable physical plausibility guidance directly within the denoising process of a task-aligned grasp diffusion model . Extensive experiments demonstrate the effectiveness of our framework.

\begin{ack}
This work received no external funding.
The authors declare no competing interests.
\end{ack}

\bibliographystyle{unsrt}
\bibliography{references}

\newpage

\appendix

\section{Further Analysis and Additional Ablation Studies}
\label{sec:app_ablation}
Below, we conduct further analysis and additional ablation studies on the standard train-test split of DexGYSNet with training following DexGYS~\cite{wei2024graspasyousay}, unless otherwise specified.

\noindent\textbf{Additional analysis on the gradient-free optimal control law.}
In the main paper, we propose the NDPP-Grasp framework, which derives a gradient-free optimal control law to enforce non-differentiable physical plausibility constraints during grasp denoising. To evaluate this design, we have compared NDPP-Grasp against four variants in Tab.~\ref{tab:ablation_control} of the main paper. Below, we provide more details about these variants.
\textbf{Variant A} introduces differentiable surrogate losses for non-differentiable physical plausibility constraints, and incorporates these surrogate losses into the diffusion training objective alongside task alignment. For example, for the hand-object penetration constraint, we adopt the penetration loss in~\cite{wei2024graspasyousay}, which approximates the non-differentiable penetration constraint via the signed distance field of the object. At inference time, this variant performs standard denoising without any denoising-time guidance.
\textbf{Variant B} uses the same surrogate losses as Variant A, but applies them at inference time rather than during training. Specifically, it adopts existing denoising-time guidance techniques~\cite{song2023loss} that leverage the gradients of these surrogate losses to compute the guidance term at each denoising step.
\textbf{Variant C} approximates the gradients of the non-differentiable constraint evaluators via zeroth-order finite-difference approximation~\cite{spall2002multivariate}. At each denoising step, it samples random perturbation vectors, queries the constraint evaluator on the corresponding terminal grasps, and aggregates the evaluations into a finite-difference gradient estimate that is used as the guidance term.
\textbf{Variant D} applies reinforcement learning~\cite{williams1992simple} to derive a guidance term from the non-differentiable constraint evaluator. At each denoising step, it draws noise increments, treats the negative physical plausibility cost as a reward, and updates the guidance term along the estimated policy-gradient direction.

Moreover, we further extend this analysis with three additional variants that all operate during the denoising process, and report the results in Tab.~\ref{tab:ablation_control_supp}.
\textbf{Variant E} adopts the evolution strategy~\cite{salimans2017evolution} to derive a guidance term from the non-differentiable constraint evaluator. At each denoising step, it draws a population of noise increments, evaluates each on the corresponding terminal grasp, and aggregates the evaluations into a population-level finite-difference estimate as the guidance term.
\textbf{Variant F} trains a separate binary classifier for each non-differentiable physical plausibility constraint to predict its satisfaction on a grasp, and uses the gradient of each classifier as the guidance term during denoising. Note that, unlike NDPP-Grasp, this variant requires additional classifier training.
\textbf{Variant G} draws $M$ candidate noise increments at each denoising step, evaluates each on the corresponding terminal grasp, and directly selects the candidate that yields the best evaluation as the noise increment of this step, without computing any guidance term.
As shown in Tab.~\ref{tab:ablation_control_supp}, NDPP-Grasp consistently outperforms all variants, further demonstrating the effectiveness of our gradient-free optimal control law for leveraging non-differentiable physical plausibility constraints during grasp denoising.

\begin{table}[h]
\centering
\caption{Evaluation of our derived gradient-free optimal control law.}
\resizebox{0.8\textwidth}{!}{
\begin{tabular}{l|ccc|ccc|ccc}
\toprule
\multirow{2}{*}{Method} & \multicolumn{3}{c|}{Physical Qual.} & \multicolumn{3}{c|}{Task Align.} & \multicolumn{3}{c}{Diversity} \\
 & Succ. $\uparrow$ & $Q_1$ $\uparrow$ & Pen. $\downarrow$ & FID $\downarrow$ & CD $\downarrow$ & Con. $\downarrow$ & $\delta_t$ $\uparrow$ & $\delta_r$ $\uparrow$ & $\delta_q$ $\uparrow$ \\
\midrule
Variant A & 54.27\% & 0.071 & 0.378 & 12.64 & 1.987 & 0.061 & 4.213 & 41.28 & 4.317 \\
Variant B & 62.91\% & 0.084 & 0.211 & 6.583 & 1.192 & 0.036 & 6.094 & 55.21 & 6.092 \\
Variant C & 60.74\% & 0.079 & 0.238 & 6.728 & 1.203 & 0.038 & 6.024 & 54.41 & 6.038 \\
Variant D & 59.86\% & 0.077 & 0.247 & 6.793 & 1.214 & 0.039 & 5.978 & 53.97 & 5.991 \\
Variant E & 61.32\% & 0.080 & 0.231 & 6.706 & 1.213 & 0.038 & 6.057 & 54.83 & 6.061 \\
Variant F & 62.45\% & 0.082 & 0.221 & 6.687 & 1.205 & 0.037 & 6.072 & 54.96 & 6.078 \\
Variant G & 64.74\% & 0.085 & 0.208 & 6.642 & 1.198 & 0.038 & 6.076 & 53.42 & 5.892 \\
\midrule
Ours & \textbf{67.05\%} & \textbf{0.088} & \textbf{0.182} & \textbf{6.487} & \textbf{1.184} & \textbf{0.035} & \textbf{6.132} & \textbf{55.74} & \textbf{6.127} \\
\bottomrule
\end{tabular}
}
\label{tab:ablation_control_supp}
\end{table}

\noindent\textbf{Impact of the lookahead horizon $H$.} 
In our framework, the lookahead horizon $H$ controls how many future denoising steps the amortized lookahead strategy uses to predict the terminal grasp via Tweedie's formula when estimating the optimal control law (see Sec.~\ref{sec:lookahead}). We set $H$ to 20, and here further examine the effect of different choices of $H$, ranging from 1 to 50. The results are summarized in Tab.~\ref{tab:ablation_H}. As shown, the performance of our framework improves notably as $H$ increases from 1 to 20, while further increasing $H$ from 20 to 50 yields saturated gains. We therefore set $H = 20$ in our framework.
\begin{table}[h]
\centering
\caption{Evaluation of the lookahead horizon $H$.}
\resizebox{0.8\textwidth}{!}{
\begin{tabular}{l|ccc|ccc|ccc}
\toprule
\multirow{2}{*}{Method} & \multicolumn{3}{c|}{Physical Qual.} & \multicolumn{3}{c|}{Task Align.} & \multicolumn{3}{c}{Diversity} \\
 & Succ. $\uparrow$ & $Q_1$ $\uparrow$ & Pen. $\downarrow$ & FID $\downarrow$ & CD $\downarrow$ & Con. $\downarrow$ & $\delta_t$ $\uparrow$ & $\delta_r$ $\uparrow$ & $\delta_q$ $\uparrow$ \\
\midrule
$H = 1$  & 61.92\% & 0.080 & 0.218 & 6.892 & 1.247 & 0.040 & 5.974 & 54.21 & 5.967 \\
$H = 10$ & 65.42\% & 0.086 & 0.197 & 6.642 & 1.213 & 0.037 & 6.072 & 55.04 & 6.068 \\
$H = 20$ & 67.05\% & 0.088 & 0.182 & 6.487 & 1.184 & 0.035 & 6.132 & 55.74 & 6.127 \\
$H = 30$ & 67.06\% & 0.088 & 0.182 & 6.486 & 1.184 & 0.035 & 6.132 & 55.74 & 6.127 \\
$H = 50$ & 67.07\% & 0.089 & 0.183 & 6.485 & 1.184 & 0.035 & 6.133 & 55.74 & 6.127 \\
\bottomrule
\end{tabular}
}
\label{tab:ablation_H}
\end{table}

\noindent\textbf{Impact of non-differentiable physical plausibility constraints.} 
In the main paper, we evaluate NDPP-Grasp with three non-differentiable physical plausibility constraints. To further verify the effectiveness of NDPP-Grasp, we consider three additional non-differentiable physical plausibility constraints and conduct experiments with different numbers of active constraints $K$. The three additional constraints are as follows. The \textbf{contact sufficiency} constraint encourages a sufficient number of fingertips to contact the object, with a smoothed contact-distance objective as its differentiable surrogate. The \textbf{force-closure} constraint encourages the generated grasp to satisfy force closure, with a continuous force-closure energy as its differentiable surrogate. The \textbf{simulation-based stability} constraint encourages the generated grasp to remain stable under physical perturbations, with a contact-aware stability objective~\cite{wang2023dexgraspnet} as its differentiable surrogate. In the first setting, we consider $K = 4$ constraints, comprising the three from the main paper plus the contact sufficiency constraint. In the second setting, we consider $K = 5$ constraints, further adding the force-closure constraint. In the third setting, we consider $K = 6$ constraints, further adding the simulation-based stability constraint. As shown in Tab.~\ref{tab:ablation_constraints}, NDPP-Grasp consistently improves over the DexGYS backbone across different numbers of active constraints, demonstrating its effectiveness under varying numbers of non-differentiable constraints.

\begin{table}[h]
\centering
\caption{Evaluation of NDPP-Grasp under different numbers of non-differentiable constraints.}
\resizebox{0.8\textwidth}{!}{
\begin{tabular}{c|c|ccc|ccc|ccc}
\toprule
\multirow{2}{*}{Number of constraints} & \multirow{2}{*}{Method} & \multicolumn{3}{c|}{Physical Qual.} & \multicolumn{3}{c|}{Task Align.} & \multicolumn{3}{c}{Diversity} \\
 & & Succ. $\uparrow$ & $Q_1$ $\uparrow$ & Pen. $\downarrow$ & FID $\downarrow$ & CD $\downarrow$ & Con. $\downarrow$ & $\delta_t$ $\uparrow$ & $\delta_r$ $\uparrow$ & $\delta_q$ $\uparrow$ \\
\midrule
\multirow{2}{*}{$K = 3$} & DexGYS~\cite{wei2024graspasyousay}                              & 63.31\% & 0.083 & 0.223 & 6.538 & 1.198 & 0.036 & 6.118 & 55.68 & 6.118 \\
                          & Ours (training following~\cite{wei2024graspasyousay})    & 67.05\% & 0.088 & 0.182 & 6.487 & 1.184 & 0.035 & 6.132 & 55.74 & 6.127 \\
\midrule
\multirow{2}{*}{$K = 4$} & DexGYS~\cite{wei2024graspasyousay}                              & 63.28\% & 0.083 & 0.225 & 6.541 & 1.200 & 0.037 & 6.117 & 55.67 & 6.117 \\
                          & Ours (training following~\cite{wei2024graspasyousay})    & 67.42\% & 0.088 & 0.179 & 6.485 & 1.184 & 0.035 & 6.134 & 55.75 & 6.128 \\
\midrule
\multirow{2}{*}{$K = 5$} & DexGYS~\cite{wei2024graspasyousay}                              & 63.25\% & 0.082 & 0.227 & 6.544 & 1.202 & 0.037 & 6.116 & 55.66 & 6.116 \\
                          & Ours (training following~\cite{wei2024graspasyousay})    & 67.71\% & 0.089 & 0.176 & 6.483 & 1.183 & 0.035 & 6.136 & 55.76 & 6.129 \\
\midrule
\multirow{2}{*}{$K = 6$} & DexGYS~\cite{wei2024graspasyousay}                              & 63.26\% & 0.082 & 0.230 & 6.547 & 1.204 & 0.038 & 6.115 & 55.65 & 6.115 \\
                          & Ours (training following~\cite{wei2024graspasyousay})    & 67.96\% & 0.089 & 0.174 & 6.481 & 1.183 & 0.034 & 6.137 & 55.77 & 6.130 \\
\bottomrule
\end{tabular}
}
\label{tab:ablation_constraints}
\end{table}

\noindent\textbf{Additional analysis on flow matching.} 
In the main paper, we develop NDPP-Grasp under the diffusion formulation. To further validate the broader applicability of NDPP-Grasp to other diffusion-related generative paradigms, we adapt it to AffordDexGrasp~\cite{wei2025afford}, a flow-matching-based dexterous grasp generation method that generates grasps by deterministic ODE integration with a learned velocity field $v_\theta(\mathbf{x}_t, t \mid c)$. Our adaptation proceeds as follows. \textbf{(1)} We construct a stochastic counterpart of the flow-matching ODE in the form $d\mathbf{x}_t = \tilde{v}_\theta(\mathbf{x}_t, t \mid c)\,dt + \sigma(t)\,d\mathbf{w}_t$ with a corrected drift $\tilde{v}_\theta = v_\theta + \frac{\sigma^2(t)}{2}s_\theta$, where $\sigma(t)$ is a noise schedule and $s_\theta$ is the score function, following the connection between deterministic and stochastic flow-matching sampling established in stochastic interpolant theory~\cite{albergo2025stochastic, albergo2022building}. In the ideal continuous-time limit with exact velocity and score fields, this stochastic counterpart preserves the same marginal path as the original flow-matching ODE. \textbf{(2)} Under the linear-interpolant convention, we obtain a score estimator from the learned velocity field as $s_\theta(\mathbf{x}_t, t \mid c) = -(\mathbf{x}_t - t \cdot v_\theta) / (1-t)$, which avoids retraining the velocity model. \textbf{(3)} We apply the gradient-free optimal control law and the amortized lookahead strategy of NDPP-Grasp on the resulting stochastic dynamics, with the terminal grasp predicted via the flow-matching endpoint estimate $\hat{\mathbf{x}}_1 = \mathbf{x}_t + (1-t)v_\theta$. As shown in Tab.~\ref{tab:ablation_flow_matching}, NDPP-Grasp consistently improves the physical plausibility of the generated grasps over the AffordDexGrasp baseline while maintaining the task alignment and diversity, demonstrating the broader applicability of NDPP-Grasp to flow-matching-based grasp generation.

\begin{table}[h]
\centering
\caption{Evaluation of NDPP-Grasp on the flow-matching-based AffordDexGrasp.}
\resizebox{0.8\textwidth}{!}{
\begin{tabular}{l|ccc|ccc|ccc|c}
\toprule
\multirow{2}{*}{Method} & \multicolumn{3}{c|}{Physical Qual.} & \multicolumn{3}{c|}{Task Align.} & \multicolumn{3}{c|}{Diversity} & \multirow{2}{*}{\shortstack{Time (ms) $\downarrow$}} \\
 & Succ. $\uparrow$ & $Q_1$ $\uparrow$ & Pen. $\downarrow$ & FID $\downarrow$ & CD $\downarrow$ & Con. $\downarrow$ & $\delta_t$ $\uparrow$ & $\delta_r$ $\uparrow$ & $\delta_q$ $\uparrow$ & \\
\midrule
AffordDexGrasp~\cite{wei2025afford} & 62.81\% & 0.082 & 0.228 & 6.538 & 1.198 & 0.039 & 5.924 & 53.86 & 5.918 & 493.1 \\
\textbf{Ours} (training following~\cite{wei2025afford}) & \textbf{65.34\%} & \textbf{0.087} & \textbf{0.201} & \textbf{6.502} & \textbf{1.189} & \textbf{0.037} & \textbf{5.951} & \textbf{54.07} & \textbf{5.943} & \textbf{28.4} \\
\bottomrule
\end{tabular}
}
\label{tab:ablation_flow_matching}
\end{table}

\noindent\textbf{Impact of the sampling number $M$.}
In our framework, the sampling number $M$ specifies how many uncontrolled trajectories the amortized lookahead strategy uses for Monte Carlo estimation of the optimal control law (see Sec.~\ref{sec:lookahead}). We set $M$ to 8, and here further examine the effect of different choices of $M$, ranging from 1 to 32. The results are summarized in Tab.~\ref{tab:ablation_M}. As shown, the performance of our framework improves notably as $M$ increases from 1 to 8, while further increasing $M$ from 8 to 32 yields saturated gains with inference time continuing to grow. We therefore set $M = 8$ in our framework to balance performance and computational cost.
\begin{table}[h]
\centering
\caption{Evaluation of the sampling number $M$.}
\resizebox{0.8\textwidth}{!}{
\begin{tabular}{l|ccc|ccc|ccc|c}
\toprule
\multirow{2}{*}{Method} & \multicolumn{3}{c|}{Physical Qual.} & \multicolumn{3}{c|}{Task Align.} & \multicolumn{3}{c|}{Diversity} & \multirow{2}{*}{\shortstack{Time (ms) $\downarrow$}} \\
 & Succ. $\uparrow$ & $Q_1$ $\uparrow$ & Pen. $\downarrow$ & FID $\downarrow$ & CD $\downarrow$ & Con. $\downarrow$ & $\delta_t$ $\uparrow$ & $\delta_r$ $\uparrow$ & $\delta_q$ $\uparrow$ & \\
\midrule
$M = 1$  & 63.71\% & 0.083 & 0.220 & 6.948 & 1.262 & 0.041 & 5.918 & 53.87 & 5.912 & 8.9 \\
$M = 4$  & 65.18\% & 0.086 & 0.198 & 6.658 & 1.217 & 0.037 & 6.064 & 55.02 & 6.058 & 10.1 \\
$M = 8$  & 67.05\% & 0.088 & 0.182 & 6.487 & 1.184 & 0.035 & 6.132 & 55.74 & 6.127 & 11.6 \\
$M = 16$ & 67.05\% & 0.088 & 0.182 & 6.486 & 1.184 & 0.035 & 6.132 & 55.74 & 6.127 & 14.7 \\
$M = 32$ & 67.07\% & 0.089 & 0.182 & 6.485 & 1.184 & 0.035 & 6.133 & 55.74 & 6.127 & 21.0 \\
\bottomrule
\end{tabular}
}
\label{tab:ablation_M}
\end{table}

\noindent\textbf{Additional analysis on task-agnostic dexterous grasp generation.}
In this paper, we propose the NDPP-Grasp framework for task-oriented dexterous grasp generation. Given that diverse diffusion-based methods also exist for task-agnostic dexterous grasp generation, here we further validate the broader applicability of NDPP-Grasp by integrating it into two representative task-agnostic methods, SceneDiffuser~\cite{huang2023scenediffuser} and DexDiffuser~\cite{weng2024dexdiffuser}. We evaluate the integrated methods on the DexGraspNet~\cite{wang2023dexgraspnet} dataset, following the standard task-agnostic dexterous grasping train-test split of~\cite{huang2023scenediffuser}. Specifically, we report the Success Rate under all six axis-aligned external force directions, the Success Rate under at least one direction, the Maximum Penetration Depth between the hand and the object, and the Diversity measured as the mean standard deviation of pose parameters across successful grasps. As shown in Tab.~\ref{tab:ablation_task_agnostic}, NDPP-Grasp consistently improves the physical plausibility of the generated grasps across both task-agnostic methods while maintaining the diversity, demonstrating the broader applicability of NDPP-Grasp.

\begin{table}[h]
\centering
\caption{Evaluation on task-agnostic dexterous grasp generation methods.}
\resizebox{0.8\textwidth}{!}{
\begin{tabular}{l|cccc}
\toprule
Method & \shortstack{Success Rate \\ (all directions) $\uparrow$} & \shortstack{Success Rate \\ (any direction) $\uparrow$} & \shortstack{Maximum Penetration \\ (mm) $\downarrow$} & Diversity $\uparrow$ \\
\midrule
SceneDiffuser~\cite{huang2023scenediffuser} & 26.6 & 66.9 & 31.0 & 0.15 \\
\textbf{Ours} (training following~\cite{huang2023scenediffuser}) & \textbf{31.4} & \textbf{70.1} & \textbf{22.7} & \textbf{0.16} \\
\midrule
DexDiffuser~\cite{weng2024dexdiffuser} & 38.2 & 75.4 & 27.8 & 0.13 \\
\textbf{Ours} (training following~\cite{weng2024dexdiffuser}) & \textbf{41.7} & \textbf{79.0} & \textbf{20.4} & \textbf{0.14} \\
\bottomrule
\end{tabular}
}
\label{tab:ablation_task_agnostic}
\end{table}

\noindent\textbf{Impact of $H$ in $H$-step trajectory resampling.}
In the main paper, we discussed that the \textbf{$H$-step trajectory resampling} (i.e., the natural finite-lookahead instantiation~\cite{song2023loss, shen2024understanding, efron2011tweedie}) suffers from a key tension between effectiveness and practicality. To verify the effectiveness of our method, we conduct experiments comparing our method against $H$-step trajectory resampling with different choices of $H$ from 1 to 50. As shown in Tab.~\ref{tab:ablation_H_resampling}, our method achieves comparable or better physical-plausibility quality than this method at all $H$ while consistently requiring substantially low inference time, demonstrating the effectiveness of our method which can resolve the tension.

\begin{table}[h]
\centering
\caption{Evaluation of $H$ in $H$-step trajectory resampling.}
\resizebox{0.8\textwidth}{!}{
\begin{tabular}{l|ccc|ccc|ccc|c}
\toprule
\multirow{2}{*}{Method} & \multicolumn{3}{c|}{Physical Qual.} & \multicolumn{3}{c|}{Task Align.} & \multicolumn{3}{c|}{Diversity} & \multirow{2}{*}{\shortstack{Time \\(ms) $\downarrow$}} \\
 & Succ. $\uparrow$ & $Q_1$ $\uparrow$ & Pen. $\downarrow$ & FID $\downarrow$ & CD $\downarrow$ & Con. $\downarrow$ & $\delta_t$ $\uparrow$ & $\delta_r$ $\uparrow$ & $\delta_q$ $\uparrow$ & \\
\midrule
$H$-step trajectory resampling ($H = 1$)  & 61.92\% & 0.080 & 0.218 & 6.892 & 1.247 & 0.040 & 5.974 & 54.21 & 5.967 & 10.7 \\
$H$-step trajectory resampling ($H = 10$) & 65.42\% & 0.086 & 0.197 & 6.642 & 1.213 & 0.037 & 6.072 & 55.04 & 6.068 & 58.5 \\
$H$-step trajectory resampling ($H = 20$) & 67.05\% & 0.088 & 0.182 & 6.488 & 1.184 & 0.035 & 6.131 & 55.73 & 6.126 & 109.8 \\
$H$-step trajectory resampling ($H = 30$) & 67.05\% & 0.088 & 0.182 & 6.486 & 1.184 & 0.035 & 6.132 & 55.74 & 6.127 & 164.7 \\
$H$-step trajectory resampling ($H = 50$) & 67.06\% & 0.089 & 0.183 & 6.485 & 1.184 & 0.035 & 6.133 & 55.74 & 6.127 & 274.5 \\
\midrule
Ours & 67.05\% & 0.088 & 0.182 & 6.487 & 1.184 & 0.035 & 6.132 & 55.74 & 6.127 & 11.6 \\
\bottomrule
\end{tabular}
}
\label{tab:ablation_H_resampling}
\end{table}

\noindent\textbf{Impact of the correction weight.}
In the main paper, we noted that directly reusing stored future states across denoising steps would correspond to using an incorrect future trajectory distribution, and we derived a closed-form correction weight $\rho_{\tau+1}^{(m)}$ to address this mismatch. To verify the effectiveness of this correction, we test a variant \textbf{Reuse without correction} that sets $\rho_{\tau+1}^{(m)} = 1$ to directly reuse stored future states without correction. As shown in Tab.~\ref{tab:ablation_correction}, our method outperforms this variant, demonstrating the effectiveness of the correction weight for valid reuse of stored trajectories across denoising steps in our method.

\begin{table}[h]
\centering
\caption{Evaluation of the correction weight.}
\resizebox{0.8\textwidth}{!}{
\begin{tabular}{l|ccc|ccc|ccc}
\toprule
\multirow{2}{*}{Method} & \multicolumn{3}{c|}{Physical Qual.} & \multicolumn{3}{c|}{Task Align.} & \multicolumn{3}{c}{Diversity} \\
 & Succ. $\uparrow$ & $Q_1$ $\uparrow$ & Pen. $\downarrow$ & FID $\downarrow$ & CD $\downarrow$ & Con. $\downarrow$ & $\delta_t$ $\uparrow$ & $\delta_r$ $\uparrow$ & $\delta_q$ $\uparrow$ \\
\midrule
Reuse without correction & 51.43\% & 0.064 & 0.413 & 9.187 & 1.582 & 0.058 & 5.421 & 51.23 & 5.402 \\
\midrule
Ours & 67.05\% & 0.088 & 0.182 & 6.487 & 1.184 & 0.035 & 6.132 & 55.74 & 6.127 \\
\bottomrule
\end{tabular}
}
\label{tab:ablation_correction}
\end{table}

\noindent\textbf{User study.}
To further evaluate NDPP-Grasp from a human perspective, we conduct a user study comparing NDPP-Grasp against the DexGYS. For each method, we generate 100 grasps, resulting in a total of 200 grasps for evaluation. We then recruit 10 participants for the user study. For each grasp, we render it as an interactive 3D visualization in HTML, allowing participants to freely rotate and zoom in their browser to inspect the grasp from arbitrary viewpoints. Participants are presented with the rendered grasps from both methods in random order to eliminate potential bias, together with the corresponding task instruction. Participants are then asked to score each grasp on a scale of 1 to 5 according to two evaluation criteria, including the physical plausibility and the task alignment. The average scores for each method are reported in Tab.~\ref{tab:user_study}. As shown, NDPP-Grasp consistently outperforms the DexGYS on both criteria, demonstrating its effectiveness from a human perspective.

\begin{table}[h]
\centering
\caption{User study comparing NDPP-Grasp against the DexGYS baseline.}
\resizebox{0.5\textwidth}{!}{
\begin{tabular}{l|cc}
\toprule
Method & Physical Plausibility $\uparrow$ & Task Alignment $\uparrow$ \\
\midrule
DexGYS~\cite{wei2024graspasyousay} & 3.2 & 3.8 \\
\textbf{Ours} (training following~\cite{wei2024graspasyousay}) & \textbf{4.2} & \textbf{4.0} \\
\bottomrule
\end{tabular}
}
\label{tab:user_study}
\end{table}

\section{Additional Qualitative Results}
\label{sec:app_qualitative}
\begin{figure}[h]
\centering
\includegraphics[width=\linewidth]{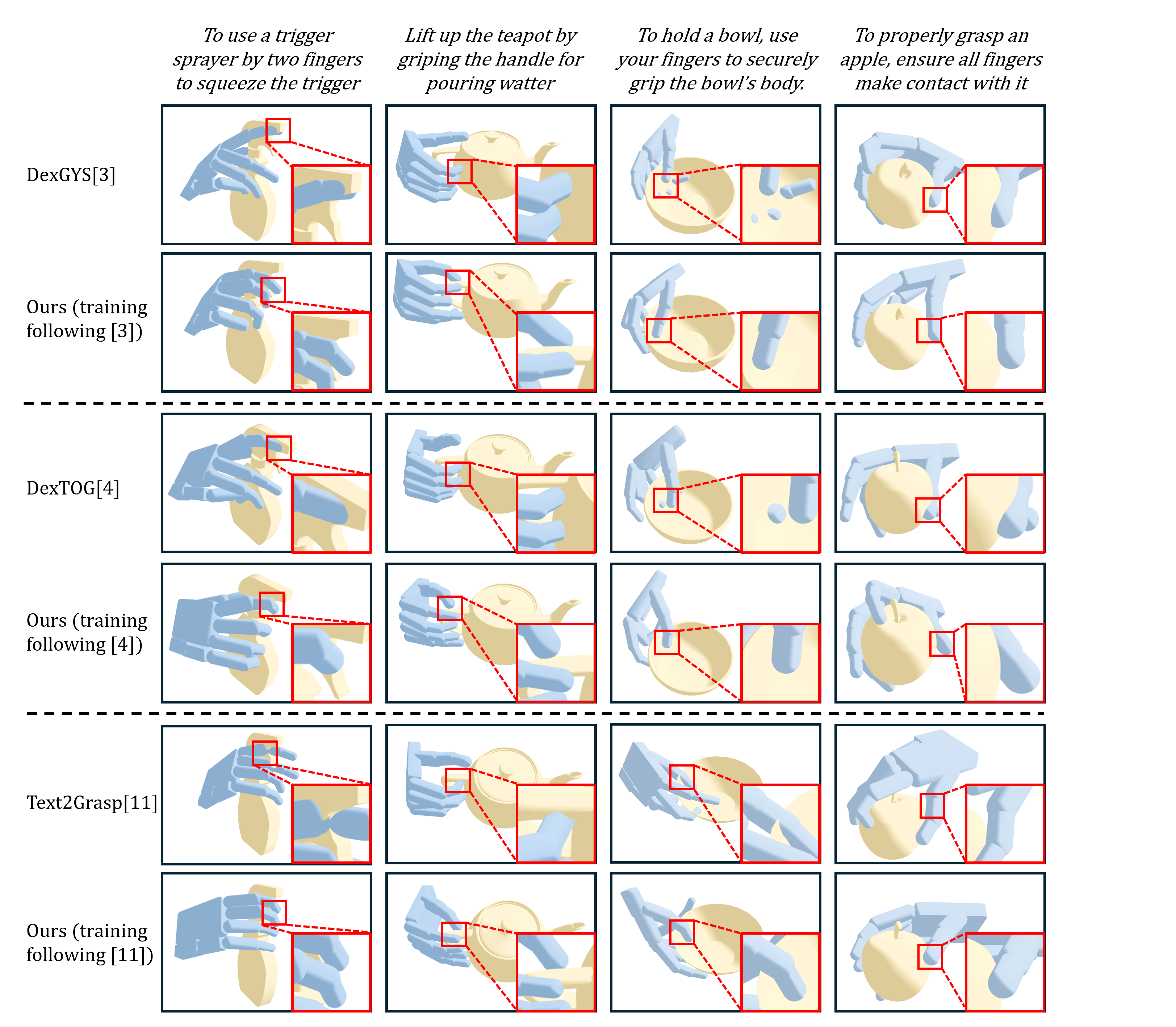}
\caption{Additional qualitative comparisons between our method and two additional baselines, DexTOG and Text2Grasp, on DexGYSNet. Key contact regions are zoomed in \textcolor{red}{red boxes}. }
\label{fig:qualitative_supp}
\end{figure}

In the main paper, we provide qualitative comparisons between our method and DexGYS~\cite{wei2024graspasyousay} in Fig.~\ref{fig:qualitative}. Here, we further provide qualitative comparisons between our method and two additional existing methods, DexTOG~\cite{zhang2024dextog} and Text2Grasp~\cite{chang2025text2grasp}, in Fig.~\ref{fig:qualitative_supp}. As shown, our method consistently produces grasps that better satisfy physical plausibility constraints while preserving task alignment and grasp diversity, while both existing methods often produce grasps with physical infeasibilities.

\section{Real-world Experiments}
\label{sec:app_realworld}

\noindent\textbf{Experimental setup.}

Geometric perception provides spatial information for robotic decision-making~\cite{wei2024graspasyousay,xiang2024overlapmamba}. We follow the real-world experimental setup of~\cite{wei2024graspasyousay} and adopt their DexGYS model as our backbone, using a six-degree-of-freedom robotic arm equipped with a dexterous hand and an RGB-D camera for visual perception. Specifically, the dexterous hand we use is the OmniHand Lingdong 2025, an anthropomorphic five-finger hand with 16 degrees of freedom (4 for the thumb and 3 for each of the remaining four fingers), of which 10 are actively driven. 
We collect several daily life objects, including both in-distribution objects from the DexGYSNet dataset and out-of-distribution objects. For each object-task pair, we conduct 10 grasping trials and report the success rate, where a trial is counted as successful if the dexterous hand stably grasps the object in a task-aligned configuration.

\noindent\textbf{Experiment Pipeline.}
Following~\cite{wei2024graspasyousay}, we first obtain the partial point cloud of the target object via visual grounding and SAM~\cite{kirillov2023segment}, and then complete it into a full point cloud through a point cloud completion network as the input to our framework. 
We deploy the ShadowHand-parameterized output via a pre-defined kinematic mapping from ShadowHand joints to the OmniHand's 10 actuated DoFs, while inactive joints follow their default coupling.
During execution, we first move the robotic arm to the pose of the predicted dexterous hand root node, and then drive the joints of the dexterous hand to the predicted pose to complete the grasp.

\begin{figure}[t]
\centering
\includegraphics[width=\linewidth]{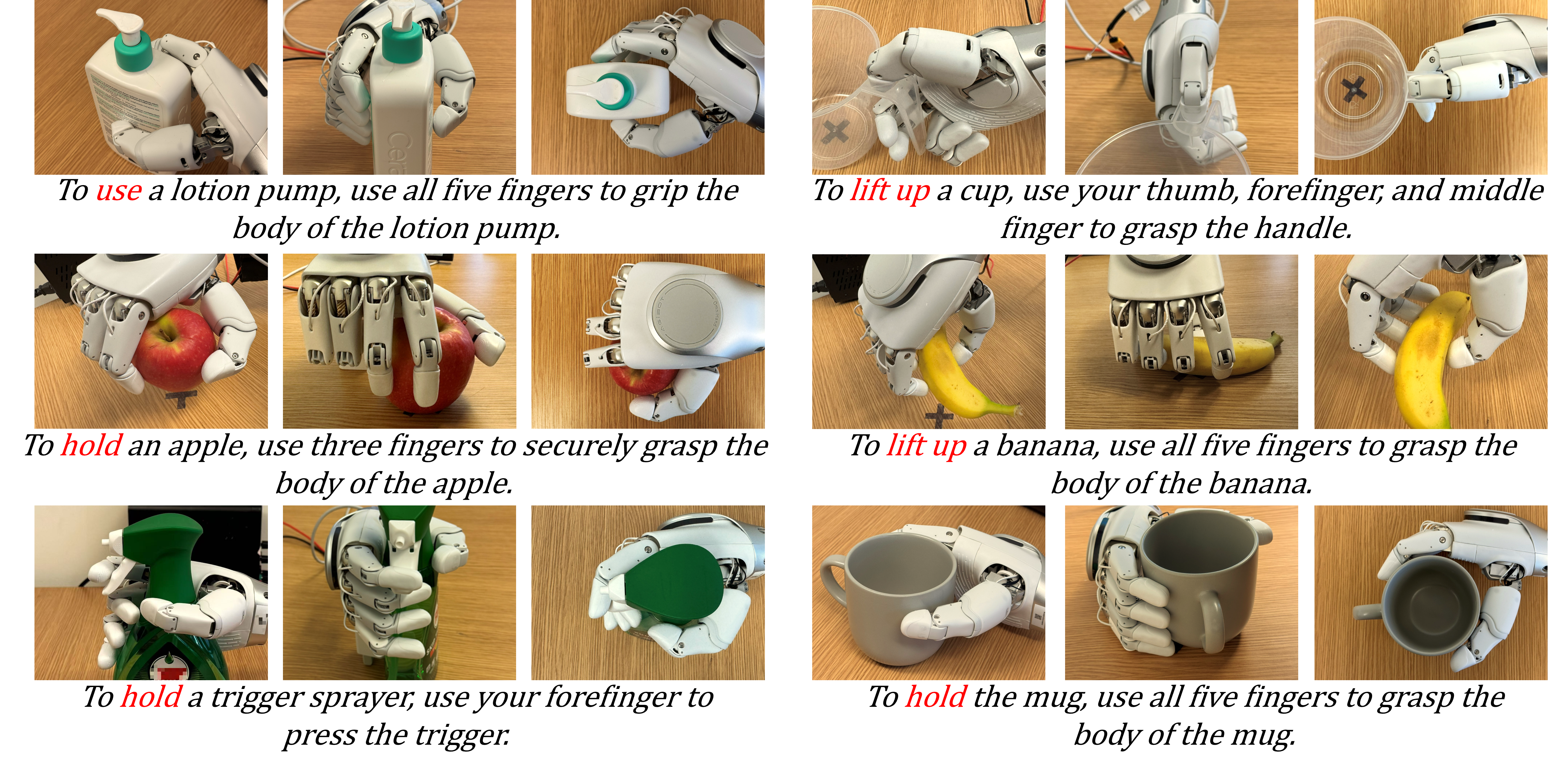}
\caption{Real-world grasping results of NDPP-Grasp on a variety of daily life objects, including objects from both in-distribution and out-of-distribution categories with respect to DexGYSNet. Across all objects, our method drives the dexterous hand to perform task-aligned and physically plausible task-oriented grasps in the real world.}
\label{fig:realworld}
\end{figure}

\noindent\textbf{Experimental results.}
As shown in Fig.~\ref{fig:realworld}, our method enables the dexterous hand to perform task-oriented grasping in a task-aligned and physically plausible manner across various objects. Quantitatively, we report per-object success rates in Tab.~\ref{tab:realworld}. Our method achieves an average success rate of 73.3\% across all object-task pairs, demonstrating the effectiveness of our method in real-world deployment.

\begin{table}[t]
\centering
\caption{Real-world grasping success rates of our method across six daily life objects.}
\label{tab:realworld}
\begin{tabular}{c|cccccc}
\toprule
Objects &Lotion pump & Cup & Apple & Banana & Trigger sprayer & Mug \\
\midrule
Success rate &7/10 & 6/10 & 9/10 & 8/10 & 6/10 & 8/10 \\
\bottomrule
\end{tabular}
\end{table}

\section{More Details about Experiments}
\label{sec:app_experiments}

\noindent\textbf{Datasets.}
We evaluate NDPP-Grasp on two widely adopted dexterous grasping benchmarks: DexGYSNet~\cite{wei2024graspasyousay} and DexTOG-80K~\cite{zhang2024dextog}. DexGYSNet is a language-guided dexterous grasping dataset comprising 50,000 ShadowHand grasps on 1,800 common household objects, where each grasp is paired with fine-grained natural language guidance describing how the hand interacts with the object. DexTOG-80K is a task-oriented dexterous grasping dataset comprising 80,000 ShadowHand grasps on 80 objects across five manipulation tasks (stapler clicking, spray bottle pressing, spray bottle triggering, bottle cap twisting, and ballpoint pen pressing), where each task is described in natural language. 

\noindent\textbf{The open-set evaluation on DexGYSNet.}
To more comprehensively evaluate the generalization ability of NDPP-Grasp beyond the standard train-test split, we follow the open-set split strategy of~\cite{wei2025afford}, originally introduced for their open-set table-top dexterous grasping benchmark, and apply it to DexGYSNet. Specifically, we first partition all object categories in DexGYSNet into seen and unseen categories. Then, 80\% of the objects from the seen categories are used for training, while the remaining 20\% of the seen-category objects together with all objects from the unseen categories form the test set. This setup jointly evaluates two complementary aspects of generalization: in-category generalization to unseen instances within seen categories, and cross-category generalization to entirely unseen categories.

\noindent\textbf{Metrics.}
We provide more details on the metrics used to evaluate NDPP-Grasp in our main paper, organized by the three aspects of physical quality, task alignment, and diversity. For physical quality, the \textbf{success rate (Succ.)} on DexGYSNet is evaluated in the Isaac Gym simulator, where each generated grasp is subjected to gravity from six different directions, and a grasp is counted as successful if it can withstand at least one direction~\cite{wei2024graspasyousay}. The \textbf{overall success rate (Overall-Succ.)} on DexTOG-80K is evaluated by executing each generated grasp with a learned task-execution policy in simulation; a grasp is counted as successful only if the policy completes the corresponding manipulation task with the grasp~\cite{zhang2024dextog}. The \textbf{$Q_1$} metric measures the norm of the smallest external wrench that disrupts grasp stability, with the contact threshold set to 1\,cm and the penetration threshold set to 5\,mm; grasps with maximal penetration depth above 5\,mm are deemed invalid and assigned $Q_1 = 0$. The \textbf{maximal penetration depth (Pen.)} reports the largest penetration distance from the object point cloud to the generated hand mesh in centimeters. The \textbf{collision-free rate ($\eta_f$)} on DexTOG-80K reports the proportion of generated grasps with no hand-object penetration. For task alignment on DexGYSNet, the \textbf{Fréchet Inception Distance (FID)} measures the distribution-level distance between generated and ground-truth grasps using point cloud features; the \textbf{Chamfer Distance (CD)} measures the chamfer distance between the generated and ground-truth hand point clouds; and the \textbf{Contact Distance (Con.)} measures the $L_2$ distance between the generated and ground-truth object-side contact maps. For diversity on DexGYSNet, $\delta_t$, $\delta_r$, and $\delta_q$ report the standard deviations of grasp translation, rotation (in Euler angles, in degrees), and joint angles across eight grasps generated under the same object and language guidance~\cite{wei2024graspasyousay}.

\noindent\textbf{Implementation details of compared methods.}
In our main paper, following DexGYS~\cite{wei2024graspasyousay} and DexTOG~\cite{zhang2024dextog}, we compare NDPP-Grasp against several task-agnostic dexterous grasp methods that do not natively support language conditioning. Below we provide more details on how these task-agnostic methods are adapted on each dataset. On DexGYSNet, following~\cite{wei2024graspasyousay}, we reimplement CVAE~\cite{sohn2015learning}, GraspTTA~\cite{jiang2021hand}, SceneDiffuser~\cite{huang2023scenediffuser}, and DGTR~\cite{xu2024dexgrasptrans} using the same point cloud encoder, training loss, and CLIP language encoder as DexGYS~\cite{wei2024graspasyousay}. The CLIP language feature is then concatenated with the object feature and the latent or noise input of each backbone to inject language conditioning, and GraspTTA further employs its test-time adaptation on top of the conditioned CVAE. On DexTOG-80K, following~\cite{zhang2024dextog}, we adapt GraspTTA~\cite{jiang2021hand} from its original CVAE-based grasp-type control to language-conditioned generation by replacing the CVAE latent with the text embedding of the task description, while keeping the rest of its training setting unchanged. The remaining compared methods (AffordDexGrasp~\cite{wei2025afford}, DexTOG~\cite{zhang2024dextog}, Text2Grasp~\cite{chang2025text2grasp}, and DexGYS~\cite{wei2024graspasyousay}) natively support language conditioning, and we train them on each dataset following the details in their original paper.

\noindent\textbf{Details of the physical plausibility constraints.}
In our main paper, following~\cite{wei2024graspasyousay, wei2025afford}, we restore the original non-differentiable physical plausibility constraints from the differentiable surrogates used in existing methods, including hand-object non-penetration, self-collision avoidance, and joint-limit constraints. Below we provide more details on each. The \textbf{hand-object non-penetration constraint} requires that no point on the hand mesh lies inside the object volume, evaluated via a hard mesh-based intersection check between the hand mesh and the object mesh, in contrast to the smooth signed-distance-based penetration losses adopted as differentiable surrogates in existing methods~\cite{wei2024graspasyousay, wei2025afford}. The \textbf{self-collision avoidance constraint} requires that no two distinct finger links of the dexterous hand intersect with each other, evaluated via pairwise collision checks across all finger-link pairs. The \textbf{joint-limit constraint} requires that every joint angle of the dexterous hand lies within its mechanically valid range, evaluated via per-joint range checks against the joint-limit specification of the dexterous hand model.
In addition to the above non-differentiable physical plausibility constraints, following AffordDexGrasp~\cite{wei2025afford}, we additionally consider two differentiable constraints based on our contact-map representation: the contact attraction loss, which pulls hand contact candidates toward the contact regions on the object surface, and the contact fingertip loss, which encourages the fingertips to remain within the contact regions. The contact map is computed at each denoising step from the current grasp configuration and object geometry alone.

\noindent\textbf{Implementation details of NDPP-Grasp.}
To obtain the predicted terminal grasp $\hat{\mathbf{g}}_T^{(m)}$ at the lookahead endpoint, we apply Tweedie's formula~\cite{efron2011tweedie} on each backbone's denoising output, with the parameterization-specific recovery formulas in Eq.~\ref{eq:app_score_y0} and Eq.~\ref{eq:app_score_eps}. The mapping $\Phi$ that converts the grasp pose $\mathcal{G}$ to the Euclidean diffusion representation $\mathbf{y}$ is implemented per backbone following each backbone's grasp parameterization, with Euler angles for the wrist rotation $r$, Euclidean coordinates for the wrist translation $t$ scaled to $[-1, 1]$, and ShadowHand joint angles for $q$ normalized via the corresponding joint-limit ranges. We handle differentiable constraints following the standard procedure in~\cite{song2023loss}. For the per-grasp inference time reported in Tab.~\ref{tab:dexgysnet} and Tab.~\ref{tab:dextog} of the main paper, we measure the average inference time over the entire test set, and repeat the measurement three times to report the mean.

\section{Theoretical Analysis about NDPP-Grasp}
\label{sec:app_theory}

\subsection{Derivation of the control-affine stochastic dynamical system in Eq.~\ref{eq:control_system} of the main paper}
\label{sec:app_dynamics}

To derive the control-affine system in Eq.~\ref{eq:control_system} of the main paper, we start from the guided reverse-time SDE in Eq.~\ref{eq:guided_sde}:
\begin{equation}
\mathrm{d}\mathbf{y}_s = -\beta(s)\Big[\tfrac{1}{2}\mathbf{y}_s + \nabla_{\mathbf{y}_s}\log p_s(\mathbf{y}_s \mid \mathcal{O}, \mathcal{L}) + \boldsymbol{\eta}_s(\mathbf{y}_s, s)\Big]\,\mathrm{d}s + \sqrt{\beta(s)}\,\mathrm{d}\bar{\mathbf{w}}_s.
\label{eq:app_guided_sde}
\end{equation}

We introduce the forward-time variable $\tau = T - s$ and let $\mathbf{g}_\tau = \mathbf{y}_{T-\tau}$. Under this change of variable, the drift sign flips since $\mathrm{d}s = -\mathrm{d}\tau$, the reverse-time Wiener increment $\sqrt{\beta(s)}\,\mathrm{d}\bar{\mathbf{w}}_s$ becomes the forward-time increment $\sqrt{\beta(T-\tau)}\,\mathrm{d}\mathbf{w}_\tau$ in distribution, and substituting $s = T-\tau$ into all time-dependent quantities of Eq.~\ref{eq:app_guided_sde} yields
\begin{equation}
\mathrm{d}\mathbf{g}_\tau = \beta(T-\tau)\Big[\tfrac{1}{2}\mathbf{g}_\tau + \nabla_{\mathbf{g}_\tau}\log p_{T-\tau}(\mathbf{g}_\tau \mid \mathcal{O}, \mathcal{L}) + \boldsymbol{\eta}_{T-\tau}(\mathbf{g}_\tau, T-\tau)\Big]\mathrm{d}\tau + \sqrt{\beta(T-\tau)}\,\mathrm{d}\mathbf{w}_\tau.
\label{eq:app_forward_sde}
\end{equation}

Recall the definitions $\mathbf{F}(\mathbf{g}_\tau, \tau) = \tfrac{1}{2}\beta(T-\tau)\mathbf{g}_\tau + \beta(T-\tau)\nabla_{\mathbf{g}_\tau}\log p_{T-\tau}(\mathbf{g}_\tau \mid \mathcal{O}, \mathcal{L})$, $\sigma(\tau) = \sqrt{\beta(T-\tau)}$, and $\mathbf{u}_\tau(\mathbf{g}_\tau, \tau) = \sqrt{\beta(T-\tau)}\,\boldsymbol{\eta}_{T-\tau}(\mathbf{g}_\tau, T-\tau)$ from Eq.~\ref{eq:control_system} of the main paper. Regrouping Eq.~\ref{eq:app_forward_sde} according to these definitions yields the control-affine stochastic dynamical system in Eq.~\ref{eq:control_system} of the main paper:
\begin{equation}
\mathrm{d}\mathbf{g}_\tau = \mathbf{F}(\mathbf{g}_\tau, \tau)\,\mathrm{d}\tau + \sigma(\tau)\big(\mathbf{u}_\tau(\mathbf{g}_\tau, \tau)\,\mathrm{d}\tau + \mathrm{d}\mathbf{w}_\tau\big).
\label{eq:app_control_affine}
\end{equation}

\subsection{Derivation of the gradient-free optimal control law in Eq.~\ref{eq:forward_control} of the main paper}
\label{sec:app_gradfree_control}

To derive the gradient-free optimal control law in Eq.~\ref{eq:forward_control} of the main paper, we start from the stochastic optimal control problem in Eq.~\ref{eq:cost} of the main paper, restated here as
\begin{equation}
\mathbf{u}^*_{\tau:T} = \arg\min_{\mathbf{u}_{\tau:T}} \mathbb{E}_{\mathcal{P}^u}\!\left[\int_\tau^T \tfrac{1}{2}\|\mathbf{u}_\nu\|^2\,\mathrm{d}\nu + f_c(\mathbf{g}_T) \,\Big|\, \mathbf{g}_\tau\right],
\label{eq:app_cost_restate}
\end{equation}
where $f_c(\mathbf{g}_T)$ is the terminal cost induced by the constraint evaluator that penalizes physical plausibility violation of the final grasp, and $\tfrac{1}{2}\|\mathbf{u}_\nu\|^2$ is the running cost that penalizes the magnitude of the control signal. Eq.~\ref{eq:app_cost_restate} requires searching over the entire sequence of control signals from time $\tau$ to $T$, which is intractable to solve directly. To make the problem solvable, we first transform it into a problem of finding a function over the state space rather than over control trajectories.

\noindent\textbf{Probabilistic representation via the desirability function.} To this end, we proceed by first introducing the value function and formulating the partial differential equation (PDE) it satisfies, then linearizing this PDE via the Hopf-Cole transformation, and finally applying the Feynman-Kac formula to obtain the desired probabilistic representation of the optimal control law.

\uline{Step (1).} To carry out this transformation, we introduce the value function, defined as the minimum expected future cost achievable from state $\mathbf{g}_\tau$ at time $\tau$:
\begin{equation}
V(\mathbf{g}_\tau, \tau) := \inf_{\mathbf{u}_{\tau:T}} \mathbb{E}_{\mathcal{P}^u}\!\left[\int_\tau^T \tfrac{1}{2}\|\mathbf{u}_\nu\|^2\,\mathrm{d}\nu + f_c(\mathbf{g}_T) \,\Big|\, \mathbf{g}_\tau\right].
\label{eq:app_value_def}
\end{equation}
By summarizing all future costs into a function of the current state alone, $V$ enables the optimal control to be obtained pointwise at each state, rather than searched over the entire control trajectory. Specifically, by standard stochastic optimal control theory~\cite{yong1999stochastic}, once $V$ is known, the optimal control at any state can be directly read off as the (negatively scaled) gradient of $V$:
\begin{equation}
\mathbf{u}^*_\tau(\mathbf{g}_\tau, \tau) = -\sigma(\tau)\,\nabla_{\mathbf{g}_\tau} V(\mathbf{g}_\tau, \tau).
\label{eq:app_u_via_V}
\end{equation}
Therefore, solving for $\mathbf{u}^*_\tau$ reduces to solving for $V$. To solve for $V$, we apply the dynamic programming principle, which states that $V$ at time $\tau$ must be consistent with $V$ at the next infinitesimal time step $\tau + \mathrm{d}\tau$ once the running cost incurred during this step is taken into account. Combining this consistency condition with the controlled dynamics in Eq.~\ref{eq:control_system} of the main paper, $V$ is shown to satisfy the stochastic Hamilton-Jacobi-Bellman (HJB) equation~\cite{yong1999stochastic}:
\begin{equation}
\frac{\partial V}{\partial \tau} - \frac{1}{2}\sigma^2(\tau)(\nabla_{\mathbf{g}_\tau} V)^\top(\nabla_{\mathbf{g}_\tau} V) + (\nabla_{\mathbf{g}_\tau} V)^\top \mathbf{F}(\mathbf{g}_\tau, \tau) + \frac{1}{2}\sigma^2(\tau)\,\mathrm{Tr}(\nabla^2_{\mathbf{g}_\tau \mathbf{g}_\tau} V) = 0,
\label{eq:app_hjb}
\end{equation}
with the terminal condition $V(\mathbf{g}_T, T) = f_c(\mathbf{g}_T)$. For notational simplicity, we omit the explicit dependence on $(\mathbf{g}_\tau, \tau)$ in $V$ in Eq.~\ref{eq:app_hjb}.

\uline{Step (2).} However, Eq.~\ref{eq:app_hjb} is non-linear in $V$, due to the quadratic term $-\tfrac{1}{2}\sigma^2(\tau)(\nabla_{\mathbf{g}_\tau} V)^\top(\nabla_{\mathbf{g}_\tau} V)$. This non-linearity makes Eq.~\ref{eq:app_hjb} intractable to solve directly in our high-dimensional grasp pose space. To remove this non-linearity, we apply the Hopf-Cole transformation, which is designed to convert HJB-type equations into linear ones by changing the variable from $V$ to its exponential counterpart. Specifically, we introduce the desirability function
\begin{equation}
\varphi(\mathbf{g}_\tau, \tau) := \exp(-V(\mathbf{g}_\tau, \tau)),
\label{eq:app_phi_def}
\end{equation}
where higher values of $\varphi$ correspond to lower expected future cost, indicating that the corresponding state is more desirable. Under this transformation, the derivatives of $V$ are related to those of $\varphi$ via the chain rule:
\begin{equation}
\nabla_{\mathbf{g}_\tau} V = -\frac{\nabla_{\mathbf{g}_\tau}\varphi}{\varphi}, \quad \nabla^2_{\mathbf{g}_\tau \mathbf{g}_\tau} V = \frac{(\nabla_{\mathbf{g}_\tau}\varphi)(\nabla_{\mathbf{g}_\tau}\varphi)^\top}{\varphi^2} - \frac{\nabla^2_{\mathbf{g}_\tau \mathbf{g}_\tau}\varphi}{\varphi}, \quad \frac{\partial V}{\partial \tau} = -\frac{1}{\varphi}\frac{\partial \varphi}{\partial \tau}.
\label{eq:app_chain_rule}
\end{equation}
Substituting Eq.~\ref{eq:app_chain_rule} into Eq.~\ref{eq:app_hjb}, the previously problematic quadratic term cancels exactly with the corresponding part of the trace term, which is the design goal of the Hopf-Cole transformation. Multiplying the resulting equation by $-\varphi$, the non-linear HJB equation is converted into the linear PDE in $\varphi$:
\begin{equation}
-\frac{\partial \varphi(\mathbf{g}_\tau, \tau)}{\partial \tau} = \mathbf{F}(\mathbf{g}_\tau, \tau)^\top \nabla_{\mathbf{g}_\tau}\varphi(\mathbf{g}_\tau, \tau) + \frac{1}{2}\sigma^2(\tau)\,\mathrm{Tr}(\nabla^2_{\mathbf{g}_\tau \mathbf{g}_\tau}\varphi(\mathbf{g}_\tau, \tau)),
\label{eq:app_linear_pde}
\end{equation}
with the terminal condition $\varphi(\mathbf{g}_T, T) = \exp(-f_c(\mathbf{g}_T))$.

\uline{Step (3).} Now that the equation in $\varphi$ is linear, we can leverage classical results from stochastic analysis to obtain its solution. We observe that Eq.~\ref{eq:app_linear_pde} is exactly the backward Kolmogorov equation associated with the uncontrolled denoising dynamics $\mathcal{P}^0$, defined as Eq.~\ref{eq:control_system} of the main paper with $\mathbf{u}_\tau \equiv 0$, which describes how a function evolves backward in time under the uncontrolled process. By the Feynman-Kac formula~\cite{oksendal2013stochastic}, the solution to such an equation admits a probabilistic representation as the conditional expectation of the terminal condition under the uncontrolled dynamics:
\begin{equation}
\varphi(\mathbf{g}_\tau, \tau) = \mathbb{E}_{\mathcal{P}^0}\!\left[\exp(-f_c(\mathbf{g}_T)) \,\big|\, \mathbf{g}_\tau\right].
\label{eq:app_feynman_kac}
\end{equation}
This representation gives the desirability function a clear interpretation: it measures how likely a trajectory starting from $\mathbf{g}_\tau$ leads to a constraint-satisfying grasp under the uncontrolled denoising dynamics. Combining Eq.~\ref{eq:app_u_via_V} with $V = -\log\varphi$, the optimal control law is now expressed in terms of $\varphi$ as
\begin{equation}
\mathbf{u}^*_\tau(\mathbf{g}_\tau, \tau) = \sigma(\tau)\,\nabla_{\mathbf{g}_\tau}\log\varphi(\mathbf{g}_\tau, \tau),
\label{eq:app_u_via_phi}
\end{equation}
which replaces the abstract value function $V$ with an explicit expectation under $\mathcal{P}^0$. However, evaluating Eq.~\ref{eq:app_u_via_phi} still requires computing the gradient $\nabla_{\mathbf{g}_\tau}\log\varphi$. The remaining task, which we accomplish below, is to convert this gradient into a form that depends only on forward evaluations of $f_c$ along uncontrolled trajectories, without requiring any gradient computation on $f_c$ itself.

\noindent\textbf{Eliminating the gradient via Markov and Gaussian structures.} The probabilistic representation derived above expresses the optimal control law via $\nabla_{\mathbf{g}_\tau}\log\varphi(\mathbf{g}_\tau, \tau)$, which still contains a gradient operator. Computing this gradient via the chain rule applied to the Feynman-Kac representation in Eq.~\ref{eq:app_feynman_kac} would require differentiating $f_c$, but as described in Sec.~\ref{sec:control} of the main paper, $f_c$ is a non-differentiable constraint evaluator and therefore admits no such gradient. Our key insight, therefore, is to eliminate the gradient operator at the level of $\nabla_{\mathbf{g}_\tau}\log\varphi$ itself, so that the optimal control law can be evaluated using only forward queries of $f_c$. Specifically, we proceed by first using the Markov property of $\mathcal{P}^0$ to transfer the gradient operator from the conditioning state to the next-step state, then exploiting the Gaussian structure of the Brownian increment via Stein's lemma to convert this gradient into a weighted expectation, and finally collapsing the resulting nested expectations into a single expectation over the entire uncontrolled trajectory.

\uline{Step (1).} To begin, we transfer the gradient from the conditioning state $\mathbf{g}_\tau$ to the random next-step state $\mathbf{g}_{\tau+\mathrm{d}\tau}$. This transfer is essential because gradients with respect to a random variable can later be eliminated through its distributional structure, while gradients with respect to a fixed conditioning variable cannot. By the Markov property of $\mathcal{P}^0$ together with the tower rule of conditional expectations, $\varphi$ at the current state can be rewritten as a one-step rollout to the next-step state under $\mathcal{P}^0$:
\begin{equation}
\varphi(\mathbf{g}_\tau, \tau) = \mathbb{E}_{\mathcal{P}^0}\!\left[\varphi(\mathbf{g}_{\tau+\mathrm{d}\tau}, \tau+\mathrm{d}\tau) \,\big|\, \mathbf{g}_\tau\right],
\label{eq:app_tower_phi}
\end{equation}
where $\mathbf{g}_{\tau+\mathrm{d}\tau} = \mathbf{g}_\tau + \mathbf{F}(\mathbf{g}_\tau, \tau)\,\mathrm{d}\tau + \sigma(\tau)\,\mathrm{d}\mathbf{w}_\tau$ is obtained from one Euler-Maruyama step of $\mathcal{P}^0$ from $\mathbf{g}_\tau$. Taking the gradient on both sides of Eq.~\ref{eq:app_tower_phi} with respect to $\mathbf{g}_\tau$, and applying the standard interchange of differentiation and expectation followed by the chain rule, we obtain
\begin{equation}
\nabla_{\mathbf{g}_\tau}\varphi(\mathbf{g}_\tau, \tau) = \mathbb{E}_{\mathcal{P}^0}\!\left[\nabla_{\mathbf{g}_{\tau+\mathrm{d}\tau}}\varphi(\mathbf{g}_{\tau+\mathrm{d}\tau}, \tau+\mathrm{d}\tau) \cdot \big(\mathbf{I} + \nabla_{\mathbf{g}_\tau}\mathbf{F}(\mathbf{g}_\tau, \tau)\,\mathrm{d}\tau\big) \,\Big|\, \mathbf{g}_\tau\right],
\label{eq:app_grad_chain}
\end{equation}
since $\partial \mathbf{g}_{\tau+\mathrm{d}\tau}/\partial \mathbf{g}_\tau = \mathbf{I} + \nabla_{\mathbf{g}_\tau}\mathbf{F}(\mathbf{g}_\tau, \tau)\,\mathrm{d}\tau$. The Jacobian term $\nabla_{\mathbf{g}_\tau}\mathbf{F}\,\mathrm{d}\tau$ is of order $O(\mathrm{d}\tau)$ and vanishes in the continuous-time limit, yielding
\begin{equation}
\nabla_{\mathbf{g}_\tau}\varphi(\mathbf{g}_\tau, \tau) = \mathbb{E}_{\mathcal{P}^0}\!\left[\nabla_{\mathbf{g}_{\tau+\mathrm{d}\tau}}\varphi(\mathbf{g}_{\tau+\mathrm{d}\tau}, \tau+\mathrm{d}\tau) \,\big|\, \mathbf{g}_\tau\right].
\label{eq:app_grad_transfer}
\end{equation}
Through this step, the gradient operation has been transferred from the conditioning state to the random next-step state, setting the stage for its elimination in Step~(2).

\uline{Step (2).} Although Step~(1) has transferred the gradient from $\mathbf{g}_\tau$ to $\mathbf{g}_{\tau+\mathrm{d}\tau}$, the gradient operator itself still remains. To eliminate this operator, we exploit the Gaussian structure of the Brownian increment driving the next-step state. Specifically, we observe that $\mathbf{g}_{\tau+\mathrm{d}\tau}$ depends linearly on the Brownian increment $\mathrm{d}\mathbf{w}_\tau$, which can be standardized as $\mathrm{d}\mathbf{w}_\tau = \sqrt{\mathrm{d}\tau}\,\boldsymbol{\xi}$ with $\boldsymbol{\xi} \sim \mathcal{N}(0, \mathbf{I})$. This standardized form admits the application of Stein's lemma, which states that for any standard Gaussian variable $\boldsymbol{\xi} \sim \mathcal{N}(0, \mathbf{I})$ and any differentiable function $h(\boldsymbol{\xi})$ with bounded gradient,
\begin{equation}
\mathbb{E}\!\left[\boldsymbol{\xi}\,h(\boldsymbol{\xi})\right] = \mathbb{E}\!\left[\nabla_{\boldsymbol{\xi}}h(\boldsymbol{\xi})\right].
\label{eq:app_stein}
\end{equation}
Stein's lemma is exactly the tool we need: it converts the expectation of a gradient (right-hand side) into an expectation weighted by the Gaussian variable itself (left-hand side), thereby eliminating the gradient operator entirely.

To apply Stein's lemma, we set $h(\boldsymbol{\xi}) := \varphi(\mathbf{g}_{\tau+\mathrm{d}\tau}, \tau+\mathrm{d}\tau)$ where $\mathbf{g}_{\tau+\mathrm{d}\tau} = \mathbf{g}_\tau + \mathbf{F}(\mathbf{g}_\tau, \tau)\,\mathrm{d}\tau + \sigma(\tau)\sqrt{\mathrm{d}\tau}\,\boldsymbol{\xi}$. By the chain rule, $\nabla_{\boldsymbol{\xi}}h = \nabla_{\mathbf{g}_{\tau+\mathrm{d}\tau}}\varphi \cdot \sigma(\tau)\sqrt{\mathrm{d}\tau}$. Substituting into Eq.~\ref{eq:app_stein} and rearranging gives
\begin{equation}
\mathbb{E}_{\mathcal{P}^0}\!\left[\nabla_{\mathbf{g}_{\tau+\mathrm{d}\tau}}\varphi(\mathbf{g}_{\tau+\mathrm{d}\tau}, \tau+\mathrm{d}\tau) \,\big|\, \mathbf{g}_\tau\right] = \frac{1}{\sigma(\tau)\sqrt{\mathrm{d}\tau}}\,\mathbb{E}_{\mathcal{P}^0}\!\left[\boldsymbol{\xi}\,\varphi(\mathbf{g}_{\tau+\mathrm{d}\tau}, \tau+\mathrm{d}\tau) \,\big|\, \mathbf{g}_\tau\right].
\label{eq:app_stein_applied}
\end{equation}
Combining Eq.~\ref{eq:app_stein_applied} with Eq.~\ref{eq:app_grad_transfer} from Step~(1) and recalling $\mathrm{d}\mathbf{w}_\tau = \sqrt{\mathrm{d}\tau}\,\boldsymbol{\xi}$, we obtain
\begin{equation}
\sigma(\tau)\,\nabla_{\mathbf{g}_\tau}\varphi(\mathbf{g}_\tau, \tau)\,\mathrm{d}\tau = \mathbb{E}_{\mathcal{P}^0}\!\left[\varphi(\mathbf{g}_{\tau+\mathrm{d}\tau}, \tau+\mathrm{d}\tau)\,\mathrm{d}\mathbf{w}_\tau \,\big|\, \mathbf{g}_\tau\right].
\label{eq:app_grad_to_dw}
\end{equation}
The gradient operator has now been completely eliminated: the right-hand side involves only an expectation of $\varphi$ at the next-step state, weighted by the Brownian increment $\mathrm{d}\mathbf{w}_\tau$.

\uline{Step (3).} Eq.~\ref{eq:app_grad_to_dw} still contains the desirability function $\varphi(\mathbf{g}_{\tau+\mathrm{d}\tau}, \tau+\mathrm{d}\tau)$ at the next-step state, which is itself defined as an expectation under $\mathcal{P}^0$ and therefore cannot be directly evaluated by forward computation on $f_c$. To obtain a final expression that depends only on the terminal cost $f_c$, we expand this inner $\varphi$ via the Feynman-Kac representation in Eq.~\ref{eq:app_feynman_kac}, applied at the next-step state:
\begin{equation}
\varphi(\mathbf{g}_{\tau+\mathrm{d}\tau}, \tau+\mathrm{d}\tau) = \mathbb{E}_{\mathcal{P}^0}\!\left[\exp(-f_c(\mathbf{g}_T)) \,\big|\, \mathbf{g}_{\tau+\mathrm{d}\tau}\right].
\label{eq:app_phi_at_next}
\end{equation}
Substituting Eq.~\ref{eq:app_phi_at_next} into Eq.~\ref{eq:app_grad_to_dw}, the right-hand side becomes a nested conditional expectation, in which the outer expectation is conditioned on $\mathbf{g}_\tau$ and the inner one is conditioned on $\mathbf{g}_{\tau+\mathrm{d}\tau}$. Since the Brownian increment $\mathrm{d}\mathbf{w}_\tau$ is determined by $(\mathbf{g}_\tau, \mathbf{g}_{\tau+\mathrm{d}\tau})$, it is measurable with respect to the inner conditioning and can therefore be absorbed into the inner expectation. The two nested conditional expectations can then be merged via the tower rule into a single expectation over the entire trajectory from $\mathbf{g}_\tau$ to $\mathbf{g}_T$ under $\mathcal{P}^0$:
\begin{equation}
\sigma(\tau)\,\nabla_{\mathbf{g}_\tau}\varphi(\mathbf{g}_\tau, \tau)\,\mathrm{d}\tau = \mathbb{E}_{\mathcal{P}^0}\!\left[\exp(-f_c(\mathbf{g}_T))\,\mathrm{d}\mathbf{w}_\tau \,\big|\, \mathbf{g}_\tau\right].
\label{eq:app_grad_final}
\end{equation}
Combining Eq.~\ref{eq:app_grad_final} with $\mathbf{u}^*_\tau = \sigma(\tau)\,\nabla_{\mathbf{g}_\tau}\log\varphi = \sigma(\tau)\,\nabla_{\mathbf{g}_\tau}\varphi/\varphi$ from Eq.~\ref{eq:app_u_via_phi}, where the denominator $\varphi(\mathbf{g}_\tau, \tau)$ takes the Feynman-Kac representation in Eq.~\ref{eq:app_feynman_kac}, we arrive at the gradient-free optimal control law in Eq.~\ref{eq:forward_control} of the main paper:
\begin{equation}
\mathbf{u}^*_\tau(\mathbf{g}_\tau, \tau)\,\mathrm{d}\tau = \frac{\mathbb{E}_{\mathcal{P}^0}\!\left[\exp(-f_c(\mathbf{g}_T))\,\mathrm{d}\mathbf{w}_\tau \,\big|\, \mathbf{g}_\tau\right]}{\mathbb{E}_{\mathcal{P}^0}\!\left[\exp(-f_c(\mathbf{g}_T)) \,\big|\, \mathbf{g}_\tau\right]}.
\label{eq:app_gradfree_final}
\end{equation}
The resulting expression involves only forward evaluations of $f_c$ along uncontrolled trajectories under $\mathcal{P}^0$, with no gradient computation on $f_c$ at any point.

\subsection{Theoretical guarantee of posterior sampling under the optimal control law}
\label{sec:app_posterior}

In Sec.~\ref{sec:control} of the main paper, we stated that the controlled denoising process under the optimal control law $\mathbf{u}^*_\tau$ samples its terminal distribution from the target posterior in Eq.~\ref{eq:posterior} of the main paper. This section provides the formal proof of this guarantee. The proof is based on a path-integral representation that connects the transition probability of the controlled process under $\mathbf{u}^*_\tau$ to that of the uncontrolled process via the desirability function, combined with marginalization over the initial Gaussian distribution. We first state the main result as Theorem~\ref{thm:posterior}, then introduce the path-integral representation as Lemma~\ref{lem:path_integral}, and finally give the proof.

\begin{theorem}[Posterior sampling under the optimal control law]
\label{thm:posterior}
Consider the controlled process in Eq.~\ref{eq:control_system} of the main paper with initialization $\mathbf{g}_0 \sim \mathcal{N}(0, \mathbf{I})$. Let $\mathcal{P}^0$ denote the corresponding uncontrolled process obtained by setting $\mathbf{u}_\tau \equiv 0$, whose terminal marginal coincides with the text-conditioned prior, i.e.,
\begin{equation}
\mathcal{P}^0(\mathbf{g}_T) = p(\mathbf{g}_T \mid \mathcal{O}, \mathcal{L}).
\label{eq:app_uncontrolled_terminal}
\end{equation}
For terminal cost $f_c(\mathbf{g}_T) = -\log p(c \mid \mathbf{g}_T) + \mathrm{const}$, the optimal control law $\mathbf{u}^*_\tau$ derived in Eq.~\ref{eq:forward_control} of the main paper induces a controlled process $\mathcal{P}^{\mathbf{u}^*}$ whose terminal marginal equals the target posterior in Eq.~\ref{eq:posterior} of the main paper, i.e.,
\begin{equation}
\mathcal{P}^{\mathbf{u}^*}(\mathbf{g}_T) = p(\mathbf{g}_T \mid \mathcal{O}, \mathcal{L}, c).
\label{eq:app_posterior_match}
\end{equation}
\end{theorem}

To prove Theorem~\ref{thm:posterior}, we first invoke a classical result from stochastic optimal control theory~\cite{yong1999stochastic}, which expresses the transition probability of the controlled process under $\mathbf{u}^*_\tau$ as the corresponding transition probability of the uncontrolled process modulated by the desirability function.

\begin{lemma}[Path-integral representation]
\label{lem:path_integral}
For any two time points $0 \leq \tau_1 < \tau_2 \leq T$, the transition probability of the controlled process under the optimal control law $\mathbf{u}^*_\tau$ from $\mathbf{g}_{\tau_1}$ to $\mathbf{g}_{\tau_2}$ satisfies
\begin{equation}
\mathcal{P}^{\mathbf{u}^*}_{\tau_1, \tau_2}(\mathbf{g}_{\tau_1}, \mathbf{g}_{\tau_2}) = \mathcal{P}^0_{\tau_1, \tau_2}(\mathbf{g}_{\tau_1}, \mathbf{g}_{\tau_2}) \cdot \frac{\varphi(\mathbf{g}_{\tau_2}, \tau_2)}{\varphi(\mathbf{g}_{\tau_1}, \tau_1)},
\label{eq:app_path_integral}
\end{equation}
where $\mathcal{P}^0_{\tau_1, \tau_2}$ denotes the corresponding transition probability of the uncontrolled process, and $\varphi$ is the desirability function defined in Eq.~\ref{eq:app_phi_def}.
\end{lemma}

The proof has four steps. We first rewrite the terminal cost in an analytically convenient form and compute the desirability function at the initial time. We next evaluate the desirability function at the terminal time. We then apply Lemma~\ref{lem:path_integral} to obtain the transition probability of the controlled process from $\tau = 0$ to $\tau = T$. We finally marginalize over the initial Gaussian distribution and apply Bayes' rule to recover the target posterior.

\uline{Step (1).} We evaluate the desirability function at the initial time. The terminal cost $f_c(\mathbf{g}_T) = -\log p(c \mid \mathbf{g}_T) + \mathrm{const}$ is unique only up to an additive constant, since adding a constant to $f_c$ leaves the optimal control law $\mathbf{u}^*_\tau$ unchanged given that $\mathbf{u}^*_\tau = \sigma(\tau) \nabla_{\mathbf{g}_\tau} \log \varphi(\mathbf{g}_\tau, \tau)$ depends only on the gradient of $\log \varphi$, which is invariant under multiplying $\varphi$ by a positive constant. We exploit this freedom to choose the additive constant so that the terminal cost takes the analytically convenient form
\begin{equation}
f_c(\mathbf{g}_T) = \log \frac{p_\mathbf{g}(\mathbf{g}_T)}{p_\mathbf{g}(\mathbf{g}_T \mid c)},
\label{eq:app_fc_form}
\end{equation}
where $p_\mathbf{g}(\mathbf{g}_T) := \mathcal{P}^0_{0, T}(\mathbf{g}, \mathbf{g}_T)$ is the terminal marginal of the uncontrolled process started from $\mathbf{g}_0 = \mathbf{g}$, and $p_\mathbf{g}(\mathbf{g}_T \mid c) := p(c \mid \mathbf{g}_T) p_\mathbf{g}(\mathbf{g}_T) / p(c)$ is the corresponding posterior at the terminal time. Here we use the fact that the initial pure noise $\mathbf{g}_0$ is independent of the constraint $c$, so $p(c \mid \mathbf{g}) = p(c)$. The equivalence of Eq.~\ref{eq:app_fc_form} with $-\log p(c \mid \mathbf{g}_T) + \mathrm{const}$ follows from $\log \frac{p_\mathbf{g}(\mathbf{g}_T)}{p_\mathbf{g}(\mathbf{g}_T \mid c)} = \log \frac{p(c)}{p(c \mid \mathbf{g}_T)} = -\log p(c \mid \mathbf{g}_T) + \log p(c)$, where $\log p(c)$ is the chosen additive constant.

With this form of $f_c$, the desirability function at the initial time admits a closed-form value. By the Feynman-Kac representation in Eq.~\ref{eq:app_feynman_kac},
\begin{equation}
\varphi(\mathbf{g}, 0) = \mathbb{E}_{\mathcal{P}^0}\!\left[\exp(-f_c(\mathbf{g}_T)) \mid \mathbf{g}_0 = \mathbf{g}\right] = \int \frac{p_\mathbf{g}(\mathbf{g}_T \mid c)}{p_\mathbf{g}(\mathbf{g}_T)}\,p_\mathbf{g}(\mathbf{g}_T)\,\mathrm{d}\mathbf{g}_T = \int p_\mathbf{g}(\mathbf{g}_T \mid c)\,\mathrm{d}\mathbf{g}_T = 1,
\label{eq:app_phi_initial}
\end{equation}
where the last equality uses the fact that $p_\mathbf{g}(\cdot \mid c)$ is a normalized probability distribution.

\uline{Step (2).} We evaluate the desirability function at the terminal time. By the boundary condition of the desirability function, $\varphi(\mathbf{g}_T, T) = \exp(-f_c(\mathbf{g}_T))$. Substituting Eq.~\ref{eq:app_fc_form},
\begin{equation}
\varphi(\mathbf{g}_T, T) = \frac{p_\mathbf{g}(\mathbf{g}_T \mid c)}{p_\mathbf{g}(\mathbf{g}_T)}.
\label{eq:app_phi_terminal}
\end{equation}

\uline{Step (3).} We now apply Lemma~\ref{lem:path_integral} with $\tau_1 = 0$ and $\tau_2 = T$. Substituting Eq.~\ref{eq:app_phi_initial} and Eq.~\ref{eq:app_phi_terminal} into Eq.~\ref{eq:app_path_integral},
\begin{equation}
\mathcal{P}^{\mathbf{u}^*}_{0, T}(\mathbf{g}, \mathbf{g}_T) = \mathcal{P}^0_{0, T}(\mathbf{g}, \mathbf{g}_T) \cdot \frac{\varphi(\mathbf{g}_T, T)}{\varphi(\mathbf{g}, 0)} = p_\mathbf{g}(\mathbf{g}_T) \cdot \frac{p_\mathbf{g}(\mathbf{g}_T \mid c) / p_\mathbf{g}(\mathbf{g}_T)}{1} = p_\mathbf{g}(\mathbf{g}_T \mid c).
\label{eq:app_transition_posterior}
\end{equation}
The transition probability of the controlled process from $\mathbf{g}_0 = \mathbf{g}$ to $\mathbf{g}_T$ is therefore exactly the posterior terminal distribution conditioned on the same initial state.

\uline{Step (4).} To obtain the unconditional terminal marginal of the controlled process, we marginalize the transition probability $\mathcal{P}^{\mathbf{u}^*}_{0, T}$ over the initial Gaussian distribution $p_0(\mathbf{g}) = \mathcal{N}(0, \mathbf{I})$. Using Eq.~\ref{eq:app_transition_posterior} and substituting the definition $p_\mathbf{g}(\mathbf{g}_T \mid c) = p(c \mid \mathbf{g}_T) p_\mathbf{g}(\mathbf{g}_T) / p(c)$,
\begin{equation}
\mathcal{P}^{\mathbf{u}^*}(\mathbf{g}_T) = \int \mathcal{P}^{\mathbf{u}^*}_{0, T}(\mathbf{g}, \mathbf{g}_T)\,p_0(\mathbf{g})\,\mathrm{d}\mathbf{g} = \int p_\mathbf{g}(\mathbf{g}_T \mid c)\,p_0(\mathbf{g})\,\mathrm{d}\mathbf{g} = \frac{p(c \mid \mathbf{g}_T)}{p(c)} \int p_\mathbf{g}(\mathbf{g}_T)\,p_0(\mathbf{g})\,\mathrm{d}\mathbf{g}.
\label{eq:app_marginal_step}
\end{equation}
By the construction of the uncontrolled process, marginalizing the conditional terminal marginals against the initial distribution recovers the text-conditioned prior, i.e.,
\begin{equation}
\int p_\mathbf{g}(\mathbf{g}_T)\,p_0(\mathbf{g})\,\mathrm{d}\mathbf{g} = \mathcal{P}^0(\mathbf{g}_T) = p(\mathbf{g}_T \mid \mathcal{O}, \mathcal{L}),
\label{eq:app_text_prior_recovery}
\end{equation}
which gives
\begin{equation}
\mathcal{P}^{\mathbf{u}^*}(\mathbf{g}_T) = \frac{p(c \mid \mathbf{g}_T)\,p(\mathbf{g}_T \mid \mathcal{O}, \mathcal{L})}{p(c)}.
\label{eq:app_marginalized}
\end{equation}
By Bayes' rule, the right-hand side of Eq.~\ref{eq:app_marginalized} equals $p(\mathbf{g}_T \mid \mathcal{O}, \mathcal{L}, c)$, the target posterior in Eq.~\ref{eq:posterior} of the main paper. This completes the proof.

\subsection{Derivation of the finite-lookahead form of the optimal control law}
\label{sec:app_lookahead}
The optimal control law in Eq.~\ref{eq:app_gradfree_final} requires expectations over full trajectories from $\mathbf{g}_\tau$ all the way to the terminal step $\mathbf{g}_T$. Simulating such full trajectories at every denoising step is prohibitively expensive in practice, and as discussed in Sec.~\ref{sec:lookahead} of the main paper, our framework instead uses an H-step lookahead with Tweedie-based terminal-grasp prediction. This section provides the formal derivation that converts the full-trajectory expectations in Eq.~\ref{eq:app_gradfree_final} into H-step trajectory expectations. 

\uline{Step (1).} To begin, we observe that the integrand in both the numerator and denominator of Eq.~\ref{eq:app_gradfree_final} depends on the trajectory beyond $\mathbf{g}_{\tau+H}$ only through the terminal state $\mathbf{g}_T$, since $\mathrm{d}\mathbf{w}_\tau$ is determined by $(\mathbf{g}_\tau, \mathbf{g}_{\tau+1})$ alone. This structure allows us to condition on the H-step lookahead trajectory and absorb everything beyond it into a single inner conditional expectation. Specifically, we denote the H-step lookahead trajectory by $\Gamma^H_\tau := (\mathbf{g}_{\tau+1}, \ldots, \mathbf{g}_{\tau+H})$, and apply the tower rule of conditional expectations together with the Markov property of $\mathcal{P}^0$. The numerator of Eq.~\ref{eq:app_gradfree_final} can then be rewritten as
\begin{equation}
\mathbb{E}_{\mathcal{P}^0}\!\left[\exp(-f_c(\mathbf{g}_T))\,\mathrm{d}\mathbf{w}_\tau \,\big|\, \mathbf{g}_\tau\right] = \mathbb{E}_{\mathcal{P}^0(\Gamma^H_\tau \mid \mathbf{g}_\tau)}\!\left[\mathbb{E}_{\mathcal{P}^0}\!\left[\exp(-f_c(\mathbf{g}_T)) \,\big|\, \mathbf{g}_{\tau+H}\right] \cdot \mathrm{d}\mathbf{w}_\tau\right],
\label{eq:app_tower_lookahead}
\end{equation}
where the outer expectation is over the H-step lookahead trajectory $\Gamma^H_\tau$ given $\mathbf{g}_\tau$, and the inner expectation is over the remaining trajectory beyond $\mathbf{g}_{\tau+H}$. By the Feynman-Kac formula in Eq.~\ref{eq:app_feynman_kac}, the inner expectation is exactly the desirability function evaluated at the lookahead endpoint:
\begin{equation}
\mathbb{E}_{\mathcal{P}^0}\!\left[\exp(-f_c(\mathbf{g}_T)) \,\big|\, \mathbf{g}_{\tau+H}\right] = \varphi(\mathbf{g}_{\tau+H}, \tau+H).
\label{eq:app_phi_at_endpoint}
\end{equation}
The same argument applies to the denominator. Substituting Eq.~\ref{eq:app_phi_at_endpoint} into Eq.~\ref{eq:app_tower_lookahead}, and applying the corresponding identity for the denominator, Eq.~\ref{eq:app_gradfree_final} reduces to
\begin{equation}
\mathbf{u}^*_\tau(\mathbf{g}_\tau, \tau)\,\mathrm{d}\tau = \frac{\mathbb{E}_{\mathcal{P}^0(\Gamma^H_\tau \mid \mathbf{g}_\tau)}\!\left[\varphi(\mathbf{g}_{\tau+H}, \tau+H)\,\mathrm{d}\mathbf{w}_\tau\right]}{\mathbb{E}_{\mathcal{P}^0(\Gamma^H_\tau \mid \mathbf{g}_\tau)}\!\left[\varphi(\mathbf{g}_{\tau+H}, \tau+H)\right]}.
\label{eq:app_lookahead_phi}
\end{equation}
The expectations are now taken over only the H-step lookahead trajectory, with the dependence on the remaining trajectory entirely absorbed into the endpoint quantity $\varphi(\mathbf{g}_{\tau+H}, \tau+H)$.

\uline{Step (2).} Eq.~\ref{eq:app_lookahead_phi} still contains the endpoint desirability function $\varphi(\mathbf{g}_{\tau+H}, \tau+H)$, which is itself an expectation over the remaining trajectory and therefore cannot be evaluated by forward computation on $f_c$. To remove this remaining expectation, we apply Tweedie's formula~\cite{efron2011tweedie} to predict the terminal grasp from the lookahead endpoint, as introduced in Sec.~\ref{sec:lookahead} of the main paper. Under the Tweedie operator $\mathcal{T}_{\tau+H}(\cdot)$ defined in the main paper, the single-step prediction of the terminal grasp from the lookahead endpoint is $\hat{\mathbf{g}}_T(\mathbf{g}_{\tau+H}) = \mathcal{T}_{\tau+H}(\mathbf{g}_{\tau+H})$, which we use to approximate the endpoint desirability function as
\begin{equation}
\varphi(\mathbf{g}_{\tau+H}, \tau+H) = \exp\!\big(-f_c(\hat{\mathbf{g}}_T(\mathbf{g}_{\tau+H}))\big).
\label{eq:app_phi_tweedie}
\end{equation}
Defining the lookahead weight as
\begin{equation}
W^H_\tau(\Gamma^H_\tau) := \exp\!\big(-f_c(\hat{\mathbf{g}}_T(\mathbf{g}_{\tau+H}))\big),
\label{eq:app_lookahead_weight}
\end{equation}
we obtain the finite-lookahead form of the optimal control law:
\begin{equation}
\mathbf{u}^*_\tau(\mathbf{g}_\tau, \tau)\,\mathrm{d}\tau = \frac{\mathbb{E}_{\mathcal{P}^0(\Gamma^H_\tau \mid \mathbf{g}_\tau)}\!\left[W^H_\tau(\Gamma^H_\tau)\,\mathrm{d}\mathbf{w}_\tau\right]}{\mathbb{E}_{\mathcal{P}^0(\Gamma^H_\tau \mid \mathbf{g}_\tau)}\!\left[W^H_\tau(\Gamma^H_\tau)\right]}.
\label{eq:app_lookahead_finite}
\end{equation}
Although $W^H_\tau$ depends on $\Gamma^H_\tau$ only through its endpoint $\mathbf{g}_{\tau+H}$ under the single-step Tweedie prediction, we deliberately write its argument as the entire trajectory $\Gamma^H_\tau$. This trajectory-level notation provides a uniform interface for the importance-sampled reuse equality derived in Sec.~\ref{sec:app_amortized}, where the proposal and target distributions differ in the entire trajectory rather than just the endpoint.

\uline{Step (3).} Both the numerator and the denominator in Eq.~\ref{eq:app_lookahead_finite} are expectations of H-step trajectory functions with respect to $\mathcal{P}^0(\Gamma^H_\tau \mid \mathbf{g}_\tau)$: the numerator integrand is $W^H_\tau(\Gamma^H_\tau) \cdot \mathrm{d}\mathbf{w}_\tau$, and the denominator integrand is $W^H_\tau(\Gamma^H_\tau)$ itself. To handle these and other such expectations uniformly in the amortized lookahead derivation in Sec.~\ref{sec:app_amortized}, we abstract them into a single generic form. Let
\begin{equation}
\Psi(\Gamma^H_\tau) = \Psi(\mathbf{g}_{\tau+1}, \ldots, \mathbf{g}_{\tau+H})
\label{eq:app_traj_functional}
\end{equation}
denote a real-valued function applied to the H-step lookahead trajectory $\Gamma^H_\tau$, used to compute the quantities entering the finite-lookahead Monte Carlo instantiation. We refer to such a $\Psi$ as a trajectory functional. Under this notation, both the numerator and the denominator in Eq.~\ref{eq:app_lookahead_finite} take the unified form $\mathbb{E}_{\mathcal{P}^0(\Gamma^H_\tau \mid \mathbf{g}_\tau)}[\Psi(\Gamma^H_\tau)]$, recovered respectively by setting $\Psi(\Gamma^H_\tau) = W^H_\tau(\Gamma^H_\tau) \cdot \mathrm{d}\mathbf{w}_\tau$ and $\Psi(\Gamma^H_\tau) = W^H_\tau(\Gamma^H_\tau)$. This unified form is exactly the form taken by the left-hand side of the reuse equality in Eq.~\ref{eq:reuse_correction} of the main paper. The corresponding right-hand side, the importance-sampled correction-weighted form that enables reuse of stored trajectories, is derived in Sec.~\ref{sec:app_amortized}.

\subsection{Derivation of the reuse equality and the closed-form correction weight}
\label{sec:app_amortized}

In the amortized lookahead procedure described in Sec.~\ref{sec:lookahead} of the main paper, the H-step lookahead segment at denoising step $\tau+1$ overlaps with the segment stored from step $\tau$ in the time indices $\tau+2$ to $\tau+H$, motivating the reuse of stored states instead of fresh sampling. However, the stored segment was generated from the candidate state $\mathbf{g}^{(m)}_{\tau+1}$ rather than the actual current state $\mathbf{g}_{\tau+1}$, so directly treating the stored states as fresh samples would correspond to using an incorrect conditional distribution. To correct this distributional mismatch, we apply importance sampling, deriving the reuse equality in Eq.~\ref{eq:reuse_correction} of the main paper together with the closed-form expression of the correction weight $\rho^{(m)}_{\tau+1}$. 

\uline{Step (1).} To convert an expectation under the target distribution into one under the proposal distribution, we invoke the standard importance sampling identity, which states that for any two distributions $p$ and $q$ with $q$ absolutely continuous with respect to $p$, and any function $f$, we have $\mathbb{E}_p[f] = \mathbb{E}_q[(p/q) \cdot f]$. In our setting, we identify the target distribution as $\mathcal{P}^0_{\tau+2:\tau+H}(\cdot \mid \mathbf{g}_{\tau+1})$, the conditional distribution of the H-step lookahead segment given the actual current state $\mathbf{g}_{\tau+1}$, and the proposal distribution as $\mathcal{P}^0_{\tau+2:\tau+H}(\cdot \mid \mathbf{g}^{(m)}_{\tau+1})$, the same distribution given the stored candidate state $\mathbf{g}^{(m)}_{\tau+1}$. We use $(\tilde{\mathbf{g}}^{(m)}_{\tau+2}, \ldots, \tilde{\mathbf{g}}^{(m)}_{\tau+H})$ to denote a sample drawn from the target distribution and $(\mathbf{g}^{(m)}_{\tau+2}, \ldots, \mathbf{g}^{(m)}_{\tau+H})$ to denote the stored sample drawn from the proposal distribution at the previous denoising step. Applying the importance sampling identity then gives
\begin{equation}
\mathbb{E}_{\mathcal{P}^0_{\tau+2:\tau+H}(\cdot \mid \mathbf{g}_{\tau+1})}\!\left[\Psi(\tilde{\mathbf{g}}^{(m)}_{\tau+2}, \ldots, \tilde{\mathbf{g}}^{(m)}_{\tau+H})\right] = \mathbb{E}_{\mathcal{P}^0_{\tau+2:\tau+H}(\cdot \mid \mathbf{g}^{(m)}_{\tau+1})}\!\left[\rho^{(m)}_{\tau+1}\,\Psi(\mathbf{g}^{(m)}_{\tau+2}, \ldots, \mathbf{g}^{(m)}_{\tau+H})\right],
\label{eq:app_reuse_eq}
\end{equation}
which is precisely Eq.~\ref{eq:reuse_correction} of the main paper. Here $\rho^{(m)}_{\tau+1}$ is the trajectory-level likelihood ratio between the target and proposal distributions:
\begin{equation}
\rho^{(m)}_{\tau+1} := \frac{\mathcal{P}^0_{\tau+2:\tau+H}(\mathbf{g}^{(m)}_{\tau+2}, \ldots, \mathbf{g}^{(m)}_{\tau+H} \mid \mathbf{g}_{\tau+1})}{\mathcal{P}^0_{\tau+2:\tau+H}(\mathbf{g}^{(m)}_{\tau+2}, \ldots, \mathbf{g}^{(m)}_{\tau+H} \mid \mathbf{g}^{(m)}_{\tau+1})}.
\label{eq:app_rho_traj}
\end{equation}
Through this step, the reuse equality has been established, with $\rho^{(m)}_{\tau+1}$ defined as a trajectory-level likelihood ratio. However, evaluating this ratio directly requires computing the joint density of the entire H-step segment under both distributions, which is computationally expensive. The next step reduces this trajectory-level ratio to a one-step transition density ratio.

\uline{Step (2).} To simplify the trajectory-level ratio in Eq.~\ref{eq:app_rho_traj}, we exploit the Markov property of $\mathcal{P}^0$, under which the joint conditional distribution factorizes as a product of one-step transition densities:
\begin{equation}
\mathcal{P}^0_{\tau+2:\tau+H}(\mathbf{g}^{(m)}_{\tau+2}, \ldots, \mathbf{g}^{(m)}_{\tau+H} \mid \mathbf{g}_{\tau+1}) = p^0(\mathbf{g}^{(m)}_{\tau+2} \mid \mathbf{g}_{\tau+1}) \prod_{k=\tau+2}^{\tau+H-1} p^0(\mathbf{g}^{(m)}_{k+1} \mid \mathbf{g}^{(m)}_k),
\label{eq:app_factorize_target}
\end{equation}
and similarly for the proposal,
\begin{equation}
\mathcal{P}^0_{\tau+2:\tau+H}(\mathbf{g}^{(m)}_{\tau+2}, \ldots, \mathbf{g}^{(m)}_{\tau+H} \mid \mathbf{g}^{(m)}_{\tau+1}) = p^0(\mathbf{g}^{(m)}_{\tau+2} \mid \mathbf{g}^{(m)}_{\tau+1}) \prod_{k=\tau+2}^{\tau+H-1} p^0(\mathbf{g}^{(m)}_{k+1} \mid \mathbf{g}^{(m)}_k),
\label{eq:app_factorize_proposal}
\end{equation}
where $p^0(\cdot \mid \cdot)$ denotes the one-step transition density of $\mathcal{P}^0$. The two factorizations differ only in the first transition density, since the conditioning state ($\mathbf{g}_{\tau+1}$ vs.~$\mathbf{g}^{(m)}_{\tau+1}$) only affects the transition into $\mathbf{g}^{(m)}_{\tau+2}$, while all subsequent transitions $p^0(\mathbf{g}^{(m)}_{k+1} \mid \mathbf{g}^{(m)}_k)$ depend only on the preceding state $\mathbf{g}^{(m)}_k$ along the trajectory and are therefore identical between the two distributions. Substituting Eq.~\ref{eq:app_factorize_target} and Eq.~\ref{eq:app_factorize_proposal} into Eq.~\ref{eq:app_rho_traj}, all subsequent transition densities cancel exactly, leaving only the first-step transition density ratio:
\begin{equation}
\rho^{(m)}_{\tau+1} = \frac{p^0(\mathbf{g}^{(m)}_{\tau+2} \mid \mathbf{g}_{\tau+1})}{p^0(\mathbf{g}^{(m)}_{\tau+2} \mid \mathbf{g}^{(m)}_{\tau+1})}.
\label{eq:app_rho_one_step}
\end{equation}
Through this step, the trajectory-level likelihood ratio has been reduced to a single one-step transition density ratio, but $p^0(\cdot \mid \cdot)$ itself remains abstract. The next step gives its closed form by exploiting the Gaussian structure of the discretized $\mathcal{P}^0$.

\uline{Step (3).} The one-step transition density $p^0(\cdot \mid \cdot)$ is determined by the Euler-Maruyama discretization of the uncontrolled dynamics $\mathcal{P}^0$, namely Eq.~\ref{eq:control_system} of the main paper with $\mathbf{u}_\tau \equiv 0$. Under this discretization, the one-step transition from a state $\mathbf{g}_s$ to the next state $\mathbf{g}_{s+1}$ is Gaussian with mean $\mathbf{g}_s + \mathbf{F}(\mathbf{g}_s, s)$ and covariance $\sigma^2(s)\,\mathbf{I}$:
\begin{equation}
p^0(\mathbf{g}_{s+1} \mid \mathbf{g}_s) = \mathcal{N}\!\big(\mathbf{g}_{s+1};\; \mathbf{g}_s + \mathbf{F}(\mathbf{g}_s, s),\; \sigma^2(s)\,\mathbf{I}\big).
\label{eq:app_gaussian_transition}
\end{equation}
Substituting Eq.~\ref{eq:app_gaussian_transition} into Eq.~\ref{eq:app_rho_one_step} for both the numerator (with $\mathbf{g}_s = \mathbf{g}_{\tau+1}$) and the denominator (with $\mathbf{g}_s = \mathbf{g}^{(m)}_{\tau+1}$), the Gaussian normalization constants are identical and cancel exactly, leaving only the exponential terms. The correction weight $\rho^{(m)}_{\tau+1}$ then admits the closed-form expression
\begin{equation}
\rho^{(m)}_{\tau+1} = \exp(-\frac{\big\|\mathbf{g}^{(m)}_{\tau+2} - \mathbf{g}_{\tau+1} - \mathbf{F}(\mathbf{g}_{\tau+1}, \tau+1)\big\|^2}{2\sigma^2(\tau+1)} + \frac{\big\|\mathbf{g}^{(m)}_{\tau+2} - \mathbf{g}^{(m)}_{\tau+1} - \mathbf{F}(\mathbf{g}^{(m)}_{\tau+1}, \tau+1)\big\|^2}{2\sigma^2(\tau+1)}).
\label{eq:app_rho_closed_form}
\end{equation}
where a larger $\rho^{(m)}_{\tau+1}$ indicates that the $m$-th stored trajectory is more representative under the current target distribution, and a smaller value indicates the opposite. When all $\rho^{(m)}_{\tau+1}$ become small, we resample fresh trajectories from $\mathbf{g}_{\tau+1}$, which is still efficient and shares the per-step cost of the initial sampling.
This closed-form expression is forward-evaluable using $\mathbf{F}$ and $\sigma$ already available from the trained diffusion model, and can be efficiently computed at each denoising step. Substituting $\rho^{(m)}_{\tau+1}$ into Eq.~\ref{eq:app_reuse_eq} yields the importance-sampled, correction-weighted estimator over the stored trajectory, which forms the basis of the amortized lookahead procedure described in Sec.~\ref{sec:lookahead} of the main paper.

\section{Overall Process of NDPP-Grasp} 
\label{sec:app_algorithm}
Our method can be flexibly integrated into various existing diffusion-based methods for task-oriented dexterous grasp generation. During training, we do not modify the training procedure of their grasp diffusion models. Instead, we follow the original methods to train a grasp diffusion model that generates task-aligned grasps conditioned on the object and task specification. During inference, however, we replace the post-generation refinement stage commonly used for physical plausibility. Rather than first generating a task-aligned grasp and then refining it, we directly apply the proposed non-differentiable physical plausibility guidance during the denoising process of the trained diffusion model. In this way, the generated grasp is guided to be both task-aligned and physically plausible throughout denoising.
The pseudo-algorithm of our NDPP-Grasp framework is summarized in Algorithm~\ref{alg:ndpp_grasp}.

\begin{algorithm}[h]
\caption{NDPP-Grasp inference}
\label{alg:ndpp_grasp}
\begin{algorithmic}[1]
\Require Trained grasp diffusion model with score $\nabla\log p_s$, object $\mathcal{O}$, task specification $\mathcal{L}$, cost evaluator $f_c$, lookahead horizon $H$, sample size $M$, total denoising steps $T$
\Ensure Generated grasp $\mathcal{G}$
\State $\mathbf{g}_0 \sim \mathcal{N}(\mathbf{0}, \mathbf{I})$
\State Sample $M$ initial $H$-step uncontrolled suffixes from $\mathbf{g}_0$ under $\mathcal{P}^0$, storing both states and Brownian increments
\For{$\tau = 0, 1, \ldots, T-1$}
    \State $H_\tau \gets \min(H, T-\tau)$ \Comment{Truncate horizon near terminal}
    \For{$m = 1, \ldots, M$}
        \State $\hat{\mathbf{g}}_T^{(m)} \gets \mathrm{Tweedie}(\mathbf{g}_{\tau+H_\tau}^{(m)})$ \Comment{Endpoint terminal grasp prediction}
        \State $W_\tau^{H,(m)} \gets \exp(-f_c(\hat{\mathbf{g}}_T^{(m)}))$ \Comment{Forward query of $f_c$, no gradient}
        \State Compute $\rho_\tau^{(m)}$ via Eq.~\ref{eq:app_rho_closed_form} \Comment{$\rho_0^{(m)} \equiv 1$ at $\tau{=}0$}
    \EndFor
    \State Compute $\mathbf{u}_\tau\,\mathrm{d}\tau$ via Eq.~\ref{eq:app_mc_amortized_estimator} using weights $\{\rho_\tau^{(m)} W_\tau^{H,(m)}\}$ and stored increments $\{\mathrm{d}\mathbf{w}_\tau^{(m)}\}$
    \State Sample fresh $\mathrm{d}\mathbf{w}_\tau^{\mathrm{act}} \sim \mathcal{N}(\mathbf{0}, \mathrm{d}\tau\,\mathbf{I})$
    \State Advance $\mathbf{g}_{\tau+1}$ from $\mathbf{g}_\tau$ via Eq.~\ref{eq:control_system} of the main paper with $\mathbf{u}_\tau$ and $\mathrm{d}\mathbf{w}_\tau^{\mathrm{act}}$
    \State Shift and extend each stored suffix by one new uncontrolled transition under $\mathcal{P}^0$
\EndFor
\State $\mathcal{G} \gets \Phi(\mathbf{g}_T)$
\State \Return $\mathcal{G}$
\end{algorithmic}
\end{algorithm}

\section{More Details about NDPP-Grasp}
\label{sec:app_details}

\subsection{Monte Carlo instantiation under natural finite-lookahead}
\label{sec:app_mc_natural}

The optimal control law in Eq.~\ref{eq:app_lookahead_finite} expresses $\mathbf{u}^*_\tau\,\mathrm{d}\tau$ as a ratio of two expectations over H-step lookahead trajectories under the uncontrolled dynamics $\mathcal{P}^0$. This subsection describes the Monte Carlo procedure used to evaluate this ratio at each denoising step in the natural finite-lookahead instantiation introduced in Sec.~\ref{sec:lookahead} of the main paper.

At denoising step $\tau$, given the current state $\mathbf{g}_\tau$, we draw $M$ independent H-step lookahead trajectories under $\mathcal{P}^0$. For $m = 1, \ldots, M$, the $m$-th trajectory $\Gamma^{H,(m)}_\tau = (\mathbf{g}^{(m)}_{\tau+1}, \ldots, \mathbf{g}^{(m)}_{\tau+H})$ is generated by Euler-Maruyama discretization of $\mathcal{P}^0$. The first step samples
\begin{equation}
\mathbf{g}^{(m)}_{\tau+1} = \mathbf{g}_\tau + \mathbf{F}(\mathbf{g}_\tau, \tau)\,\mathrm{d}\tau + \sigma(\tau)\,\mathrm{d}\mathbf{w}^{(m)}_\tau, \qquad \mathrm{d}\mathbf{w}^{(m)}_\tau \sim \mathcal{N}(0, \mathrm{d}\tau\,\mathbf{I}),
\label{eq:app_first_step_natural}
\end{equation}
and the subsequent $H-1$ steps sample $\mathbf{g}^{(m)}_{k+1}$ from $\mathbf{g}^{(m)}_k$ for $k = \tau+1, \ldots, \tau+H-1$ via the same Euler-Maruyama discretization. For each trajectory, we then apply the Tweedie operator $\mathcal{T}_{\tau+H}$ to the lookahead endpoint to obtain the predicted terminal grasp $\hat{\mathbf{g}}^{(m)}_T = \mathcal{T}_{\tau+H}(\mathbf{g}^{(m)}_{\tau+H})$, and compute the lookahead weight
\begin{equation}
W^{H,(m)}_\tau = \exp\!\big(-f_c(\hat{\mathbf{g}}^{(m)}_T)\big),
\label{eq:app_lookahead_weight_mc}
\end{equation}
which corresponds to the $m$-th sample of the trajectory functional $W^H_\tau$ defined in Eq.~\ref{eq:app_lookahead_weight}.

With the $M$ sampled lookahead weights $\{W^{H,(m)}_\tau\}_{m=1}^M$ and the corresponding first-step Brownian increments $\{\mathrm{d}\mathbf{w}^{(m)}_\tau\}_{m=1}^M$, the two expectations in Eq.~\ref{eq:app_lookahead_finite} are estimated by their empirical averages, and the resulting Monte Carlo instantiation of the optimal control law is
\begin{equation}
\mathbf{u}^*_\tau(\mathbf{g}_\tau, \tau)\,\mathrm{d}\tau = \frac{\sum_{m=1}^{M}\,W^{H,(m)}_\tau\,\mathrm{d}\mathbf{w}^{(m)}_\tau}{\sum_{m=1}^{M}\,W^{H,(m)}_\tau},
\label{eq:app_mc_natural_estimator}
\end{equation}
where the empirical-average normalization factors cancel between numerator and denominator. This estimator gives the current-step Monte Carlo instantiation of Eq.~\ref{eq:forward_control} of the main paper under the natural finite-lookahead procedure introduced in Sec.~\ref{sec:lookahead} of the main paper.

\subsection{Monte Carlo instantiation under amortized lookahead}
\label{sec:app_mc_amortized}

In the amortized lookahead procedure introduced in Sec.~\ref{sec:lookahead} of the main paper, the lookahead trajectory at each denoising step is largely reused from the previous step rather than sampled fresh. This subsection describes the Monte Carlo procedure used to evaluate the optimal control law at the current step under this reuse, building on the reuse equality and the closed-form correction weight derived in Sec.~\ref{sec:app_amortized}.

Suppose at the previous denoising step $\tau$, we have stored $M$ trajectories $\{(\mathbf{g}^{(m)}_{\tau+1}, \mathbf{g}^{(m)}_{\tau+2}, \ldots, \mathbf{g}^{(m)}_{\tau+H})\}_{m=1}^{M}$, where each trajectory was generated by Euler-Maruyama discretization of $\mathcal{P}^0$ starting from the candidate state $\mathbf{g}^{(m)}_{\tau+1}$. At the current step $\tau+1$, given the actual current state $\mathbf{g}_{\tau+1}$, we evaluate the optimal control law using these stored trajectories in place of fresh samples.

For each $m = 1, \ldots, M$, we first reuse the overlapping segment $(\mathbf{g}^{(m)}_{\tau+2}, \ldots, \mathbf{g}^{(m)}_{\tau+H})$ directly from storage, and extend the trajectory by one step under $\mathcal{P}^0$ by sampling $\mathbf{g}^{(m)}_{\tau+H+1}$ from $\mathbf{g}^{(m)}_{\tau+H}$ via Euler-Maruyama discretization. We apply the Tweedie operator $\mathcal{T}_{\tau+H+1}$ at the new endpoint to obtain the predicted terminal grasp $\hat{\mathbf{g}}^{(m)}_T = \mathcal{T}_{\tau+H+1}(\mathbf{g}^{(m)}_{\tau+H+1})$, and compute the lookahead weight at the current step, where $W^{H,(m)}_{\tau+1} = \exp\!\big(-f_c(\hat{\mathbf{g}}^{(m)}_T)\big).$

We then compute the correction weight $\rho^{(m)}_{\tau+1}$ in closed form using Eq.~\ref{eq:app_rho_closed_form}, evaluated with the actual current state $\mathbf{g}_{\tau+1}$, the stored candidate state $\mathbf{g}^{(m)}_{\tau+1}$, and the stored next-step state $\mathbf{g}^{(m)}_{\tau+2}$. Finally, we compute the first-step Brownian increment $\mathrm{d}\mathbf{w}^{(m)}_{\tau+1}$ that enters the numerator of the control law in Eq.~\ref{eq:app_lookahead_finite}. Since this control law is evaluated under the target distribution conditional on the actual current state $\mathbf{g}_{\tau+1}$, this Brownian increment combines the stored next-step state $\mathbf{g}^{(m)}_{\tau+2}$ with the actual current state $\mathbf{g}_{\tau+1}$,
\begin{equation}
\mathrm{d}\mathbf{w}^{(m)}_{\tau+1} = \frac{\mathbf{g}^{(m)}_{\tau+2} - \mathbf{g}_{\tau+1} - \mathbf{F}(\mathbf{g}_{\tau+1}, \tau+1)\,\mathrm{d}\tau}{\sigma(\tau+1)},
\label{eq:app_dw_amortized}
\end{equation}
as required by the importance sampling framework in Sec.~\ref{sec:app_amortized}.

Substituting these quantities into the reuse equality in Eq.~\ref{eq:app_reuse_eq}, the two expectations in Eq.~\ref{eq:app_lookahead_finite} (instantiated at step $\tau+1$) are estimated by their importance-sampling-weighted empirical averages over the stored trajectories, giving the Monte Carlo instantiation of the optimal control law at the current step,
\begin{equation}
\mathbf{u}^*_{\tau+1}(\mathbf{g}_{\tau+1}, \tau+1)\,\mathrm{d}\tau = \frac{\sum_{m=1}^{M}\,\rho^{(m)}_{\tau+1}\,W^{H,(m)}_{\tau+1}\,\mathrm{d}\mathbf{w}^{(m)}_{\tau+1}}{\sum_{m=1}^{M}\,\rho^{(m)}_{\tau+1}\,W^{H,(m)}_{\tau+1}},
\label{eq:app_mc_amortized_estimator}
\end{equation}
where the empirical-average normalization factors cancel between numerator and denominator. The extended trajectory $(\mathbf{g}^{(m)}_{\tau+2}, \ldots, \mathbf{g}^{(m)}_{\tau+H+1})$ is then stored for the next denoising step, where $\mathbf{g}^{(m)}_{\tau+2}$ takes over the role of the candidate state and $(\mathbf{g}^{(m)}_{\tau+3}, \ldots, \mathbf{g}^{(m)}_{\tau+H+1})$ becomes the reused overlapping segment. 
Notably, the memory cost of these stored trajectories is negligible, since the $M$ lookahead trajectories consist of low-dimensional grasp states and amount to only $\mathcal{O}(MH)$ floats per generation. 

\subsection{Score function recovery under different denoising parameterizations.}
The reverse-time SDE in Eq.~\ref{eq:reverse_sde} of the main paper requires the score function $\nabla_{\mathbf{y}_s}\log p_s(\mathbf{y}_s \mid \mathcal{O}, \mathcal{L})$ to drive the denoising trajectory. In practice, however, different diffusion-based grasp generation backbones train the network to predict different quantities. To make our framework compatible with these backbones, here we show how the score function can be recovered in closed form from the two most common denoising parameterizations, so that the NDPP-Grasp framework can be plugged in without modification to its underlying control formulation. Specifically, when the denoising network adopts the clean-sample prediction (i.e., $\mathbf{y}_0$-parameterization) and predicts the clean grasp representation $\mathbf{y}_0^\theta(\mathbf{y}_s, s)$ from the noisy input $\mathbf{y}_s$, the score function is recovered as
\begin{equation}
\nabla_{\mathbf{y}_s}\log p_s(\mathbf{y}_s \mid \mathcal{O}, \mathcal{L}) = \frac{\sqrt{\bar{\alpha}_s}\,\mathbf{y}_0^\theta(\mathbf{y}_s, s) - \mathbf{y}_s}{1 - \bar{\alpha}_s},
\label{eq:app_score_y0}
\end{equation}
where $\bar{\alpha}_s$ is the cumulative noise schedule coefficient. When the denoising network adopts the noise prediction (i.e., $\boldsymbol{\epsilon}$-parameterization) and predicts the added noise $\boldsymbol{\epsilon}_\theta(\mathbf{y}_s, s)$, the score function is recovered as
\begin{equation}
\nabla_{\mathbf{y}_s}\log p_s(\mathbf{y}_s \mid \mathcal{O}, \mathcal{L}) = -\frac{\boldsymbol{\epsilon}_\theta(\mathbf{y}_s, s)}{\sqrt{1 - \bar{\alpha}_s}}.
\label{eq:app_score_eps}
\end{equation}
Note that the two parameterizations are mathematically equivalent under the forward diffusion relationship $\mathbf{y}_s = \sqrt{\bar{\alpha}_s}\,\mathbf{y}_0 + \sqrt{1 - \bar{\alpha}_s}\,\boldsymbol{\epsilon}$, and our framework operates at the level of the score function rather than a specific network architecture. As a result, the SOC formulation in Sec.~\ref{sec:control} of the main paper and the gradient-free derivation in Sec.~\ref{sec:app_gradfree_control} apply directly under either parameterization, allowing NDPP-Grasp to be plugged into different diffusion-based backbone trained under either parameterization without modifying the underlying control formulation.

\section{Limitations and Future Work}
\label{sec:app_limitations}

 \noindent\textbf{Failure cases and sim-to-real transfer.} When grasping objects with very smooth surfaces such as mugs, although grasp generation methods can generate a valid grasp, the dexterous hand's rigid fingertips may provide insufficient friction, potentially causing slippage. Nevertheless, we observe that our method achieves a 60\% real-robot experiment success rate even in this challenging setting, outperforming 20\% achieved by DexGYS~\cite{wei2024graspasyousay}. The inherent sim-to-real gap is a common challenge in the embodied intelligence community, and handling this challenge is an important and useful future direction in this whole area.{}

 \noindent\textbf{Potential safety implications.} \textbf{(i) Physical operation risks.} During human-robot interaction, attempting to grasp objects that exceed the robotic hand's payload capacity may cause physical harm. \textbf{(ii) Adversarial risks.} Adversarial attacks on learning-based decision-making systems may cause task failure or, in robotic deployments, even dangerous actions~\cite{robey2025jailbreaking,xiang2026monitoring,xiang2026new}. Future work in this area can include (a) further improving the robotic hand's manipulation and payload capabilities and (b) designing stronger defenses to deal with potential adversarial attacks and unauthorized reuse of diffusion models~\cite{li2025diffip}.{}

\section{Licenses}
We use the DexGYSNet dataset~\cite{wei2024graspasyousay} by following the license of \href{https://huggingface.co/datasets/wyl2077/DexGYS}{here}, and use the DexTOG-80K dataset~\cite{zhang2024dextog} by following the license of \href{https://huggingface.co/datasets/robotflow/DexTOG}{here}. 

\newpage

\newpage
\section*{NeurIPS Paper Checklist}

\begin{enumerate}

\item {\bf Claims}
    \item[] Question: Do the main claims made in the abstract and introduction accurately reflect the paper's contributions and scope?
    \item[] Answer: \answerYes{}
    \item[] Justification: In both the abstract and the introduction, we have clearly mentioned the task (scope) and the contributions of this paper

\item {\bf Limitations}
    \item[] Question: Does the paper discuss the limitations of the work performed by the authors?
    \item[] Answer: \answerYes{}
    \item[] Justification: We discuss potential failure cases, sim-to-real transfer, and safety implications in Appendix~\ref{sec:app_limitations}.

\item {\bf Theory assumptions and proofs}
    \item[] Question: For each theoretical result, does the paper provide the full set of assumptions and a complete (and correct) proof?
    \item[] Answer: \answerYes{}
    \item[] Justification: This paper does include such complete and correct theoretical analysis (proof) in Appendix.

    \item {\bf Experimental result reproducibility}
    \item[] Question: Does the paper fully disclose all the information needed to reproduce the main experimental results of the paper to the extent that it affects the main claims and/or conclusions of the paper (regardless of whether the code and data are provided or not)?
    \item[] Answer: \answerYes{}
    \item[] Justification:  This paper builds its code based on the diffusion-based task-oriented dexterous generation methods. We have clearly mentioned (via descriptions and equations) how to reproduce the main experimental results upon the  existing diffusion-based task-oriented dexterous generation methods.

\item {\bf Open access to data and code}
    \item[] Question: Does the paper provide open access to the data and code, with sufficient instructions to faithfully reproduce the main experimental results, as described in supplemental material?
    \item[] Answer: \answerNo{}
    \item[] Justification: At this submission stage, we are sorry that we do not get enough approval to open-source our code.

\item {\bf Experimental setting/details}
    \item[] Question: Does the paper specify all the training and test details (e.g., data splits, hyperparameters, how they were chosen, type of optimizer) necessary to understand the results?
    \item[] Answer: \answerYes{}
    \item[] Justification: This paper has specified its data splits, its introduced hyperparameters, and other details in the Experiment section and Appendix.

\item {\bf Experiment statistical significance}
    \item[] Question: Does the paper report error bars suitably and correctly defined or other appropriate information about the statistical significance of the experiments?
    \item[] Answer: \answerNo{}
    \item[] Justification: To make a fair comparison with the existing works, we follow their experimental settings which do not including error bars.

\item {\bf Experiments compute resources}
    \item[] Question: For each experiment, does the paper provide sufficient information on the computer resources (type of compute workers, memory, time of execution) needed to reproduce the experiments?
    \item[] Answer: \answerYes{}
    \item[] Justification: We introduce the type of GPU and the time of execution in the Experiments section.
    
\item {\bf Code of ethics}
    \item[] Question: Does the research conducted in the paper conform, in every respect, with the NeurIPS Code of Ethics \url{https://neurips.cc/public/EthicsGuidelines}?
    \item[] Answer: \answerYes{}
    \item[] Justification: The authors have read through the NeurIPS Code of Ethics and carefully conform with it.

\item {\bf Broader impacts}
    \item[] Question: Does the paper discuss both potential positive societal impacts and negative societal impacts of the work performed?
    \item[] Answer: \answerNo{}
    \item[] Justification: This paper focuses on handling a crucial problem in task-oriented dexterous grasp generation. To the best of our knowledge, this work is methodological and we do not foresee direct societal impacts specific to this work.
    
\item {\bf Safeguards}
    \item[] Question: Does the paper describe safeguards that have been put in place for responsible release of data or models that have a high risk for misuse (e.g., pre-trained language models, image generators, or scraped datasets)?
    \item[] Answer: \answerNA{}
    \item[] Justification: To the best of our knowledge, this paper does not newly bring any risk for misuse and is thus not applicable to this question.

\item {\bf Licenses for existing assets}
    \item[] Question: Are the creators or original owners of assets (e.g., code, data, models), used in the paper, properly credited and are the license and terms of use explicitly mentioned and properly respected?
    \item[] Answer: \answerYes{}
    \item[] Justification: We have discussed the licenses in our Appendix.

\item {\bf New assets}
    \item[] Question: Are new assets introduced in the paper well documented and is the documentation provided alongside the assets?
    \item[] Answer: \answerNA{}
    \item[] Justification: This paper does not release new assets.

\item {\bf Crowdsourcing and research with human subjects}
    \item[] Question: For crowdsourcing experiments and research with human subjects, does the paper include the full text of instructions given to participants and screenshots, if applicable, as well as details about compensation (if any)? 
    \item[] Answer: \answerNA{}
    \item[] Justification: This paper does not involve crowdsourcing nor research with human subjects.

\item {\bf Institutional review board (IRB) approvals or equivalent for research with human subjects}
    \item[] Question: Does the paper describe potential risks incurred by study participants, whether such risks were disclosed to the subjects, and whether Institutional Review Board (IRB) approvals (or an equivalent approval/review based on the requirements of your country or institution) were obtained?
    \item[] Answer: \answerNA{}
    \item[] Justification: This paper does not involve crowdsourcing nor research with human subjects.

\item {\bf Declaration of LLM usage}
    \item[] Question: Does the paper describe the usage of LLMs if it is an important, original, or non-standard component of the core methods in this research? Note that if the LLM is used only for writing, editing, or formatting purposes and does \emph{not} impact the core methodology, scientific rigor, or originality of the research, declaration is not required.
    \item[] Answer: \answerNA{}
    \item[] Justification:  LLM is not an important, original, or non-standard component of the core methodology in this study.

\end{enumerate}

\end{document}